%% file: paper.tex
\documentclass[10pt,twocolumn,letterpaper]{article}

\usepackage{cvpr}
\usepackage{times}
\usepackage{epsfig}
\usepackage{graphicx}
\usepackage{amsmath}
\usepackage{amssymb}
\usepackage{ctable} 
\usepackage{float} 

\usepackage{array}
\newcolumntype{L}[1]{>{\raggedright\arraybackslash}p{#1}}
\newcolumntype{C}[1]{>{\centering\arraybackslash}p{#1}}
\newcolumntype{R}[1]{>{\raggedleft\arraybackslash}p{#1}}

\def\RR{\mathbb{R}} 
\def\CC{\mathbb{C}} 
\graphicspath{{fig/}} 
\def\etal{\emph{et al.~}}

\usepackage[pagebackref=true,breaklinks=true,letterpaper=true,colorlinks,bookmarks=false]{hyperref}

\cvprfinalcopy 

\begin{document}

\title{Uncertainty-Aware CNNs for Depth Completion: \text{Uncertainty from Beginning to End}}

\author{Abdelrahman Eldesokey~~~~~Michael Felsberg~~~~~Karl Holmquist~~~~~Mikael Persson\\
{\normalsize Computer Vision Laboratory, Link\"oping University, Sweden}}

\maketitle
\thispagestyle{empty}

\begin{abstract}
	The focus in deep learning research has been mostly to push the limits of  prediction accuracy. However, this was often achieved at the cost of increased complexity, raising concerns about the interpretability and the reliability of deep networks. Recently, an increasing attention has been given to untangling the complexity of deep networks and quantifying their uncertainty for different computer vision tasks. Differently, the task of depth completion has not received enough attention despite the inherent noisy nature of depth sensors. In this work, we thus focus on modeling the uncertainty of depth data in depth completion starting from the sparse noisy input all the way to the final prediction. 
	
	We propose a novel approach to identify disturbed measurements in the input by learning an input confidence estimator in a self-supervised manner based on the normalized convolutional neural networks (NCNNs). Further, we propose a probabilistic version of NCNNs that produces a statistically meaningful uncertainty measure for the final prediction. When we evaluate our approach on the KITTI dataset for depth completion, we outperform all the existing Bayesian Deep Learning approaches in terms of prediction accuracy, quality of the uncertainty measure, and the computational efficiency. Moreover, our small network with 670k parameters performs on-par with conventional approaches with millions of parameters. 
	These results give strong evidence that separating the network into parallel uncertainty and prediction streams leads to state-of-the-art performance with accurate uncertainty estimates. 

\end{abstract}

\vspace{-5mm}
\section{Introduction}
\vspace{-2mm}
\input{sec/intro.tex}

\section{Related Work}

\input{sec/related.tex}

\section{Self-supervised Input Confidence Learning} \label{sec:3}
\input{sec/nconv.tex}

\section{Probabilistic NCNNs}\label{sec:4}
\input{sec/bdl.tex}

\section{Experiments}\label{sec:exp}
\input{sec/exp.tex}

\vspace{-2mm}
\section{Conclusion}\label{sec:conc}
\vspace{-2mm}
We proposed a self-supervised approach for estimating the input confidence for sparse data based on the NCNNs. We also introduced a probabilistic version of NCNNs that enable the to output meaningful uncertainty measures. Experiments on the KITTI dataset for unguided depth completion showed that our small network with 670k parameters achieves state-of-the-art results in terms of prediction accuracy and it provides an accurate uncertainty measure. When compared against the existing probabilistic method for dense problems, our proposed approach outperforms all of them in terms of the prediction accuracy, the quality of the uncertainty measure, and the computational efficiency. Moreover, we showed that our approach can be applied to other sparse problems as well. These results demonstrate the gains from adhering to the signal/uncertainty philosophy compared to conventional black-box models.

\noindent {\bf Acknowledgments:}
This work was 
supported by 
the Wallenberg AI, Autonomous Systems and Software Program (WASP)
and
Swedish Research Council 
 grant 2018-04673. 

{\small
\bibliographystyle{ieee_fullname}
\bibliography{egbib}
}

\onecolumn
\setcounter{equation}{0}    
\setcounter{section}{0}    

\input{supplementary}

\end{document}

%% file: sec/intro.tex

The recent surge of deep neural networks (DNNs) has led to 
remarkable breakthroughs on several computer vision tasks, \eg object classification and detection \cite{touvron2019fixing,mahajan2018exploring,li2019scale,cao2019gcnet}, semantic segmentation \cite{zhu2019improving,takikawa2019gated}, and object tracking \cite{danelljan2019atom, wang2019fast}.  However, this was achieved at the cost of increased model complexity, inducing new concerns such as: how do these black-box models infer their predictions? and how certain are they about these predictions? 
Failing to address these concerns impairs the reliability of DNNs. For instance, Huang \etal~\cite{huang2019upc} showed that it is possible to fool state-of-the-art object detectors to produce false and highly certain predictions using physical and digital manipulations. Therefore, there is a compelling need for investigating interpretability and uncertainty of DNNs to be able to trust them in safety-critical environments.

\begin{figure}[t]
	\vspace*{-2mm}
	\includegraphics[width=\columnwidth]{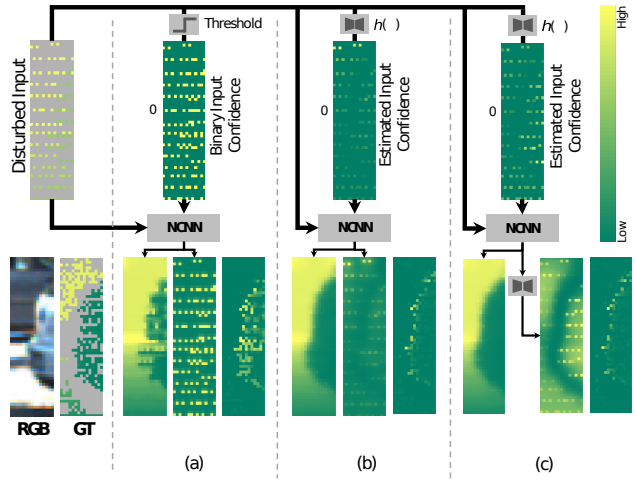}
		\vspace*{-5mm}
	\caption{The confidence $\mathbf{c}^0$ for the input data is usually unknown. NCNNs \cite{pami} assume binary input confidence, which leads to severe artifacts \textbf{(a)}. We propose to learn the input confidence in a self-supervised manner, which leads to improved prediction \textbf{(b)}. However, the output confidence $\mathbf{c}^L$ is not strongly correlated with the error $\mathbf{E}$. Therefore, we propose a probabilistic version of NCNN that produces a proper output uncertainty measure \textbf{(c)}.}
	\label{fig:intro}
		\vspace*{-3mm}
\end{figure}
 
Recently, a growing attention was given towards untangling the complexity of DNNs to enhance their reliability by analyzing how they make predictions and quantifying the uncertainty of these predictions. Probabilistic approaches such as  Bayesian deep learning (BDL) have contributed to this endeavor by modifying DNNs to output the parameters of a probabilistic distribution, \eg mean and variance, which yields uncertainty information about the predictions \cite{kendall2017uncertainties}. The availability of a reliable uncertainty measure facilitates the understanding of DNNs and applying safety procedures in case of model failure or high uncertainty. Several BDL approaches were proposed for different computer vision tasks such as object classification and segmentation \cite{gal2016dropout,lakshminarayanan2017simple,kendall2017uncertainties}, optical flow \cite{ilg2018uncertainty,gast2018lightweight}, and object detection \cite{law2018cornernet,Choi_2019_ICCV}. All these approaches assume undisturbed dense input images, but to the best of our knowledge, there exist no statistical approach that addresses sparse problems.


An essential task of this type is \emph{scene depth completion}. Modeling uncertainty for this task is crucial due to the inherent noisy and sparse nature of depth sensors, caused by multi-path interference and depth ambiguities \cite{guo2018tackling}. Previous approaches proposed to learn some intermediate confidence masks to mitigate the impact of disturbed measurements inside their networks \cite{qiu2019deeplidar,wvangansbeke_depth_2019,Xu_2019_ICCV}. 
However, none of these approaches has demonstrated the probabilistic validity of the intermediate confidence masks. Moreover, they do not provide an uncertainty measure for the final prediction. Therefore, it is still an open problem to fully model the uncertainty in DNN approaches to scene depth completion.

Gustafsson \etal \cite{gustafsson2019evaluating} made an attempt by evaluating two of the existing BDL approaches for dense regression problems, \ie MC-Dropout \cite{gal2016dropout} and ensembling \cite{lakshminarayanan2017simple}, on the task of depth completion. They utilized the Sparse-to-Dense network \cite{ma2019self} as a baseline  and modified it to estimate the parameters of a Gaussian distribution. Experiments on the KITTI-Depth dataset \cite{uhrig2017sparsity} showed that both approaches can produce high-quality uncertainty maps for the final prediction, but with the prediction accuracy severely degraded compared to the baseline model. Besides, both approaches train an ensemble of the baseline model requiring multiple inferences during test time. This leads to computational and memory overhead making these approaches unsuitable for the task of depth completion in practice due to their poor prediction accuracy and computational inefficiency.


Specifically designed for 
confidence-accompanied 
and sparse
data 
are
the normalized convolutional neural networks (NCNNs) \cite{bmvc,pami}. NCNNs consist of a serialization of confidence-equipped convolution layers that make use of an input confidence map. 
These layers produce
the output of the convolution operation as well as an output confidence that is propagated to the following layer. When applied to the problem of depth completion, input confidences at the first layer are assumed to be binary following \cite{uhrig2017sparsity}, ones at valid input points and zeros otherwise. However, this assumption is problematic since depth data can be disturbed as noted in the KITTI-Depth dataset \cite{qiu2019deeplidar}. Therefore, the use of binary masks for modeling input uncertainty in NCNNs becomes inappropriate, and hinders 
their use
as the true input confidence is \emph{unknown}. 
Also, 
the output confidence from NCNNs according to \cite{bmvc,pami} lacks any probabilistic interpretation that qualifies it as a reliable uncertainty measure.

\subsection{Contributions}
In this paper, we propose two main contributions. \emph{First}, we employ the inherent dependency of NCNNs on the input confidence to train an estimator for this confidence in a self-supervised manner. Since disturbed measurements are expected to increase the prediction error, we back-propagate the error gradients to learn the input confidence that minimizes the error. This way, the network learns to assign low confidences to disturbed measurements that increase the error and high confidences to valid measurements. This approach establishes a new methodology for handling sparse and noisy data by \emph{suppressing} the disturbed measurements before feeding them to the network. As shown empirically, this approach is more interpretable and efficient than utilizing a complex black-box model that is expected to implicitly rectify for the disturbed measurements.

\emph{Second,} we derive a probabilistic NCNN (pNCNN) framework that produces meaningful uncertainty estimates in the probabilistic sense, whereas the output confidence from the standard NCNNs lacks any probabilistic characteristics. We formulate the training process as a maximum likelihood estimation problem and we derive the loss function for pNCNN training. These reformulations are the necessary extensions for fully Bayesian NCNNs.

By applying our approach to the task of unguided depth completion on the KITTI-Depth dataset \cite{uhrig2017sparsity}, we achieve a remarkably better prediction accuracy at a very low computational cost compared to the existing BDL approaches. Moreover, the quality of the uncertainty measure from our \emph{single} network is better than BDL approaches with ensembles of 1-32 networks. When compared against non-statistical approaches, we perform on par with state-of-the-art methods with millions of parameters using a significantly smaller network (670k parameters). Besides, and contrarily to state-of-the-art methods, we produce a high-quality prediction uncertainty measure aside with the prediction. Finally, we show that our approach is applicable to other sparse problems by evaluating it on multi-path interference correction \cite{guo2018tackling} and sparse optical flow rectification.

%% file: sec/related.tex
The task of scene depth completion is receiving an increasing attention due to the impact of depth information on different computer vision tasks. Typically, it aims to produce a dense and denoised depth map $\mathbf{y}$ from a noisy sparse input $\mathbf{x}$. Several approaches were proposed to learn a mapping $\mathbf{y} = f(\mathbf{x})$ by exploiting different input modalities, where $f$ is a DNN. Ma \etal \cite{ma2019self} proposed a deep regression model that combines the sparse input depth with the corresponding RGB modality. Jaritz \etal \cite{jaritz2018sparse} evaluated different fusion schemes to combine the sparse depth with RGB images. Chen \etal \cite{learning2019yun} proposed a joint network that exploits 2D and 3D representations for the depth data. The key similarity between these approaches is that they all perform very well in terms of prediction accuracy and they implicitly handle disturbed measurements in the network. Nonetheless, none of these methods considered modeling the uncertainty of the data or the prediction.

Recently, several approaches promoted the use of confidences to filter out noisy predictions within the network. Qui \etal \cite{qiu2019deeplidar} learned confidence masks from RGB images to mask out noisy depth measurements at occluded regions. Gansbeke \etal \cite{wvangansbeke_depth_2019} proposed the use of confidences to fuse two network streams utilizing sparse depth and RGB images respectively. Similarly, Xu \etal \cite{Xu_2019_ICCV} predict a confidence mask that is used to mitigate the impact of noisy measurements on different components of their network. However, none of these methods provided any prediction uncertainty measure for the final prediction. 

This was addressed by another approach that utilizes confidences and provides an output confidence for the final prediction. Normalized convolutional neural networks (NCNNs) \cite{bmvc,pami} take sparse depth $\mathbf{x}$ and a confidence mask $\mathbf{c}^0$ as input, propagate the confidence, and produce a dense output $\mathbf{y}$ as well as an output confidence map $\mathbf{c}^L$, \ie, $(\mathbf{y}, \mathbf{c}^L) = f(\mathbf{x}, \mathbf{c}^0)$, for a DNN with $L$ layers. However, since the input confidence is unknown, a binary input confidence $\mathbf{c}^0$ is assumed, which is problematic in case of disturbed input as shown in Figure (\ref{fig:intro}a). Further, the output confidence $\mathbf{c}^L$ has no probabilistic interpretation and shows no significant correlation with the prediction error.

To address these challenges, we look at the problem from a different perspective. We propose to learn the input confidence from the disturbed measurements by employing the confidence propagation property of NCNNs. We attach a network $h$ to a NCNN and we train them end-to-end to learn the input confidence that minimizes the prediction error, \ie, $(\mathbf{y}, \mathbf{c^L}) = f(\mathbf{x}, h(\mathbf{x}))$. Further, to produce accurate uncertainty measure for the final prediction, we derive a probabilistic version of the NCNNs and we formulate the training as a maximum likelihood problem. When our proposed approach is evaluated on the KITTI-Depth dataset \cite{uhrig2017sparsity}, it performs on par with state-of-the-art approaches with millions of parameters using a significantly smaller network, while providing a highly accurate uncertainty measure for the final prediction. In contrast to BDL approaches in \cite{gustafsson2019evaluating}, we achieve excellent uncertainty estimation without sacrificing prediction accuracy or computational efficiency.

The rest of the paper is organized as follows. We briefly describe the method of NCNNs in \ref{sec:nc} and \ref{sec:ncnn}, and our proposed approach for learning the input confidence in section~\ref{sec:conf_est}. Afterwards, we introduce a probabilistic version of NCNNs, derive the loss for training, and describe our architecture in section \ref{sec:4}. Experiments and analysis are given in section \ref{sec:exp}. Finally, we conclude the paper in section~ \ref{sec:conc}.

%% file: sec/nconv.tex
The signal/confidence philosophy \cite{knutsson1993normalized} promotes the separation between the signal and its confidence for efficiently handling noisy and sparse signals. For example, this separation allows differentiating missing signal points with no information from zero-valued valid points. The normalized convolution \cite{knutsson1993normalized} is one approach that follows the this philosophy to perform the convolution operation. 

For confidence-equipped signals, the normalized convolution performs convolution using only the confident points of the signal, while estimating the non-confident ones from their vicinity using some \emph{applicability function}. This prevents noisy and missing measurements from disturbing the calculations. In this section, we give a brief description of normalized convolution and the trainable normalized convolution layer that can estimate an optimal applicability \cite{bmvc,pami}. Subsequently, we propose a novel approach to learn the input confidence in a self-supervised manner. 

Throughout the paper, we assume a global signal $\mathcal{Y}$ with a finite size $N$ that is convolved in a sliding window fashion. At each point in the signal $y_i$, a local signal $\mathbf{y}$ of size $n$ constitutes the neighborhood at this point. The local signal $\mathbf{y}$ will be referred to as \emph{the signal}, and $y_i$ will be referred to as \emph{the signal center}.

\begin{figure*}[!t]
	\vspace*{-2mm}
	\centering
	\includegraphics[width=0.95\textwidth]{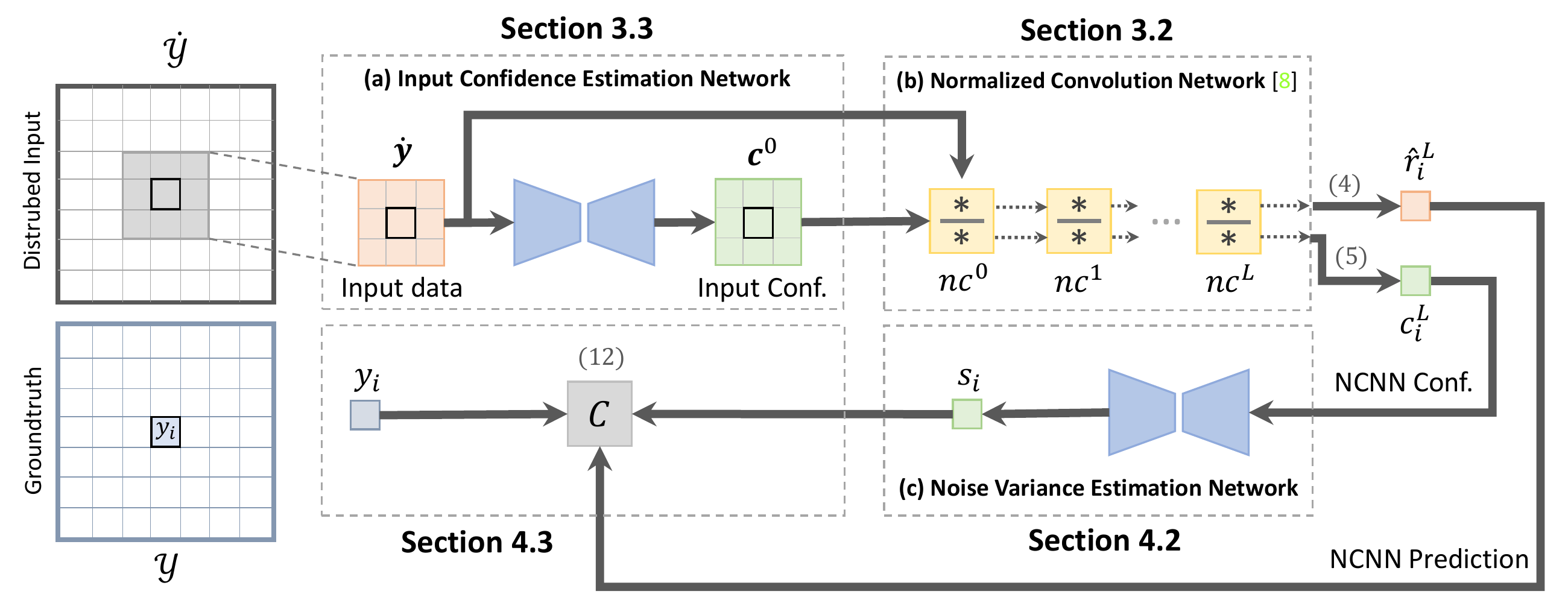}
	\caption{An overview of network architecture to predict a denoised signal $\mathcal{Y}$ from a disturbed signal $\dot{\mathcal{Y}}$. We show the pipeline for a single observation $y_i$ of the whole signal $\mathcal{Y}$. Our contributions are described in sections \ref{sec:conf_est}, \ref{sec:conf}, and \ref{sec:loss}.}
	\label{fig:arch}
		\vspace*{-2mm}
\end{figure*}


\subsection{The Normalized Convolution} \label{sec:nc}
The fundamental idea of the normalized convolution is to project the  confidence-equipped signal $\mathbf{y} \in \CC^n$ to a new subspace spanned by a set of basis functions $\{\mathbf{b}_j\}_{j=0}^m$ using only the confident parts of the signal. Afterwards, the full signal is reconstructed from this subspace, where the non-confident parts are interpolated from their vicinity using a weighting kernel denoted as the \emph{applicability function}. The confidence is provided as non-negative real vector $\mathbf{c} \in \RR_{+}^n$ that has the same length as the signal $\mathbf{y}$, while the applicability $\mathbf{a}\in \RR_{+}^n$ is usually chosen as some low-pass filter. 


If we arrange the basis functions into the columns of a matrix $\mathbf{B}$, then the image of the signal under the subspace spanned by the basis is obtained as $\mathbf{y} = \mathbf{B} \mathbf{r}$, where $\mathbf{r}$ is a vector of coordinates. These coordinates can be estimated from a weighted least-squares problem (WLS) between the signal $\mathbf{y}$ and the image of it under the new basis:
\begin{equation}\label{eq:nconv_min}
\hat{\mathbf{r}}_\text{WLS} = \arg \min_{\mathbf{r} \in \CC^m} \parallel \mathbf{B} \mathbf{r} - \mathbf{y} \parallel_\mathbf{W} \enspace,
\end{equation}
where the weights matrix $\mathbf{W}$ is a product of $\mathbf{W}_\mathbf{a} = \text{diag}(\mathbf{a})$ and ${\mathbf{W}_\mathbf{c} = \text{diag}(\mathbf{c})}$. The WLS solution is given as \cite{knutsson1993normalized}:
\begin{equation}\label{eq:wls_sol}
\hat{\mathbf{r}}_\text{WLS} = \underbrace{(\mathbf{B}^*\mathbf{W}_\mathbf{a} \mathbf{W}_\mathbf{c} \mathbf{B})^{-1}}_\text{Normalize} \underbrace{\mathbf{B}^* \mathbf{W}_\mathbf{a} \mathbf{W}_\mathbf{c} \mathbf{y}}_\text{Project} \enspace .
\end{equation}

\noindent Finally, the WLS solution $\hat{\mathbf{r}}_\text{WLS}$ can be used to approximate the signal under the new basis as:
\begin{equation}\label{eq:r_wls}
\hat{\mathbf{y}}  = \mathbf{B} \hat{\mathbf{r}}_\text{WLS} \enspace .
\end{equation}


\subsection{Normalized Convolutional Neural Networks} \label{sec:ncnn}

In normalized convolution, the applicability is chosen manually. Eldesokey \etal \cite{pami} proposed a normalized convolutional neural network layer (NCNN) that utilized the standard back-propagation in DNNs to learn the optimal applicability function $\mathbf{a}$ for a given dataset, while assuming a binary input confidence. This was achieved by using the na\"{i}ve basis in (\ref{eq:wls_sol}), \ie $\mathbf{B} = \mathbf{1}_n$:
\begin{equation}\label{eq:naive_basis}
\hat{r}_i = (\mathbf{1}_n^*\mathbf{W}_\mathbf{a} \mathbf{W}_\mathbf{c} \mathbf{1}_n)^{-1} \mathbf{1}_n^* \mathbf{W}_\mathbf{a} \mathbf{W}_\mathbf{c} \mathbf{y} = \dfrac{  \langle \mathbf{a}  | (\mathbf{y} \odot \mathbf{c}) \rangle}{\langle \mathbf{a} | \mathbf{c} \rangle},
\end{equation}
where $\mathbf{1}_n$ is a vector of ones, $\odot$ is the Hadamard product, $\langle . | . \rangle$ is the scalar product, $\hat{r}_i$ is a scalar which is equivalent to the estimated value at the signal center $\hat{y}_i$. They proposed to propagate the confidence from the NCNN layer as:
\begin{equation}\label{eq:cout}
\hat{c}_i = \dfrac{\langle \mathbf{a} | \mathbf{c} \rangle}{\langle \mathbf{1}_n | \mathbf{a} \rangle} \enspace ,
\end{equation}
where the output confidence from one layer is the input confidence to the next layer.


\subsection{Self-Supervised Input Confidence Estimation using NCNNs} \label{sec:conf_est}

The assumption of binary input confidences adopted by \cite{bmvc,pami} can be problematic in real datasets. An example is the KITTI-Depth dataset \cite{uhrig2017sparsity}, where some of the input values do not match the groundtruth due to LiDAR projection errors (shown in Figure \ref{fig:abl} top). In this case, a binary input confidence would lead to artifacts in the output as NCNNs are dependent on the input confidence as shown in the calculations of (\ref{eq:naive_basis}). This dependency of the outputs on the input confidences facilitates learning the confidences. The inclusion of the input confidences in the calculations of the output from each layer indicates that the loss of the network would constitute gradients with respect to these confidences. Therefore, we can employ these gradients to learn input confidences that minimize the loss function.

We propose to use an \emph{input confidence estimation} network that receives the input data and produces an estimate for the input confidence that is fed to the first layer of the NCNN. This network is trained end-to-end with the NCNN and the error gradients from the NCNN are back-propagated to the confidence estimation network, allowing it to learn the input confidence that minimizes the overall prediction error. We use a compact UNet \cite{unet} for the confidence estimation network with a Softplus activation at the final layer that will produce valid confidence values in the interval $[0,\infty[$. The pipeline is illustrated in Figure \ref{fig:arch} (upper part).



%% file: sec/bdl.tex
Figure (\ref{fig:intro}b) shows an example of the output confidence from the last NCNN layer when we estimate the input confidences using our proposed approach from the previous section. The figure shows that the output confidences do not exhibit a proper uncertainty measure that is strongly correlated with the error.

To obtain proper uncertainties from NCNNs, we introduce a probabilistic version of NCNNs by deriving the connection between the normalized convolution and statistical least-squares approaches. Then, we utilize this connection to produce reliable uncertainties with probabilistic characteristics. Finally, we apply the proposed theory to NCNNs and we derive a loss function for training them to produce accurate uncertainties.

\subsection{Connection between NCNN and Generalized Least-Squares}

In ordinary least-squares (OLS) problems, constant variance is assumed for all observations of the signal. Generalized least-squares (GLS), on the other hand, offers more flexibility to handle individual variance per observation. 
The \emph{weighted-least squares} problem in (\ref{eq:wls_sol}) can be viewed as a special case of the GLS, where observations are heteroskedastic with unequal noise levels. 

Assume the image of the signal under the subspace $\mathbf{B}$ is defined as ${\mathbf{y} = \mathbf{B} \mathbf{r} + \mathbf{e}}$, where $\mathbf{e}$ is a random noise variable with zero mean and variance $\sigma^2 \mathbf{V}$. This variance models the heteroscedastic uncertainty of the observations in the signal, where $\sigma^2$ is global for each signal, and $\mathbf{V}$ is a positive definite matrix describing the covariance between the observations. The GLS solution to this problem reads \cite{aitken1936iv}:

\begin{equation}\label{eq:gls_sol}
\hat{\mathbf{r}}_\text{GLS} = (\mathbf{B}^* \mathbf{V}^{-1} \mathbf{B})^{-1} \mathbf{B}^* \mathbf{V}^{-1} \mathbf{y} \enspace .
\end{equation}
When comparing the two solutions in (\ref{eq:wls_sol}) and (\ref{eq:gls_sol}), they are only equivalent if $\mathbf{V^{-1}}$ is diagonal, which leads to  $\mathbf{V}=(\mathbf{W}_\mathbf{a} \mathbf{W}_\mathbf{c})^{-1}$. The diagonality of the covariance matrix indicates that different samples of the signal are independent and have different variances depending on the confidence and the applicability function. 

We utilize the GLS solution $\hat{\mathbf{r}}_\text{GLS}$ to estimate the signal similar to (\ref{eq:r_wls}) as $\hat{\mathbf{y}}=\mathbf{B} \hat{\mathbf{r}}_\text{GLS}$. The uncertainty of $\hat{\mathbf{y}}$ can be estimated as:
\begin{equation}\label{eq:gls_cov}
\begin{split}
\text{cov}(\hat{\mathbf{y}})&= \text{cov}(\mathbf{B} \hat{\mathbf{r}}_\text{GLS}) = \mathbf{B} \ \text{cov}(\hat{\mathbf{r}}_\text{GLS}) \mathbf{B}^*  \\
& = \sigma^2 \mathbf{B}(\mathbf{B}^* \mathbf{V}^{-1} \mathbf{B})^{-1}\mathbf{B}^* \\
& = \sigma^2 \mathbf{B}(\mathbf{B}^* \mathbf{W}_\mathbf{a} \mathbf{W}_\mathbf{c}  \mathbf{B})^{-1}\mathbf{B}^* \enspace .
\end{split}
\end{equation} 

\noindent Note that $\mathbf{W}_\mathbf{a} \text{ and } \mathbf{W}_\mathbf{c}$ are non-stochastic, where the former is estimated during NCNN training and the latter can be learned using our proposed approach in section \ref{sec:conf_est}. On the other hand, $\sigma^2$ is unknown and needs to be estimated.


\subsection{Output Uncertainty for NCNNs}\label{sec:conf}

In case of NCNNs with the na\"ive basis ${\mathbf{B} = \mathbf{1}_n}$, the uncertainty measure in (\ref{eq:gls_cov}) simplifies to:
\begin{equation}\label{eq:naive_cov}
\begin{split}
\text{cov}(\hat{\mathbf{y}})&= \text{cov}(\mathbf{1}_n \hat{{r}}) = \sigma^2 \mathbf{1}_n(\mathbf{1}_n^* \mathbf{W}_\mathbf{a} \mathbf{W}_\mathbf{c}  \mathbf{1}_n)^{-1}\mathbf{1}_n^* \\ 
&= \mathbf{1}_n\dfrac{\sigma^2}{\langle \mathbf{a} | \mathbf{c} \rangle}\mathbf{1}_n^*\enspace .
\end{split}
\end{equation} 
This indicates an equal uncertainty for the whole neighborhood, but since we are only interested in signal center $\hat{y}_i$, (\ref{eq:naive_cov}) reduces to:
\begin{equation}\label{eq:naive_cov2}
\text{var}(\hat{y}_i)= \dfrac{\sigma_i^2}{\langle \mathbf{a} | \mathbf{c} \rangle}\enspace .
\end{equation} 

It is evident that the output confidence described in (\ref{eq:cout}) disregards the stochastic noise variance $\sigma_i^2$. However, to obtain a proper uncertainty measure, this variance needs to be incorporated in the output confidence. We propose to estimate the noise variance $\sigma_i^2$ from the output confidence of the last NCNN layer by means of a noise variance estimation network as illustrated in Figure \ref{fig:arch}. To achieve this, we need a loss function that allows training the proposed framework.


\subsection{The Loss Function for Probabilistic NCNNs} \label{sec:loss}
We consider each point $y_i$ in the global signal $\mathcal{Y}$, where the neighborhood at this point is the local signal $\mathbf{y}$. This local signal can be represented under some basis as $\hat{\mathbf{y}}=\mathbf{B} \hat{\mathbf{r}}$, where the estimated coordinates $\hat{\mathbf{r}}$ are calculated from (\ref{eq:gls_sol},\ref{eq:wls_sol}). We assume that the estimate of the signal follows a multivariate normal distribution ${\hat{\mathbf{y}} \sim \mathcal{N}_m (\mathbf{B} \hat{\mathbf{r}}, \sigma^2 \mathbf{B}(\mathbf{B}^* \mathbf{W}_\mathbf{a} \mathbf{W}_\mathbf{c}  \mathbf{B})^{-1}\mathbf{B}^*)}$ where the variance is defined in (\ref{eq:gls_cov}). In case of the na\"ive basis, we will have a univariate normal distribution ${\hat{y}_i \sim \mathcal{N}(\hat{r}_i, \sigma_i^2 / \langle \mathbf{a} | \mathbf{c} \rangle)}$, where the variance is defined in (\ref{eq:naive_cov2}). More formally, a NCNN outputs the mean $\hat{r}_i^L$ of the normal distribution around $\hat{y}_i$, and the scalar product $\langle \mathbf{a} | \mathbf{c} \rangle$ in the denominator of the variance. Yet, the noise variance $\sigma^2$ needs to be estimated to comply with the definition in (\ref{eq:naive_cov2}). 

We denote the variance term as $s_i = \sigma_i^2 / \langle \mathbf{a} | \mathbf{c} \rangle$, where $\mathbf{a} \text{ and } \mathbf{c}$ are the applicability and the output confidence from the \emph{last} NCNN layer. The least squares solution in (\ref{eq:naive_basis}) can be formulated as a maximum likelihood problem of a Gaussian error model for the last NCNN layer $L$: 
\begin{equation}\label{eq:likelihood}
l(\mathbf{w}) = \dfrac{1}{\sqrt{2 \pi s_i}} \exp \left(- \dfrac{ \parallel y_i - \hat{r}_i^L \parallel^2}{2 s_i} \right) \enspace ,
\end{equation}
where $\mathbf{w}$ denotes the network parameters, and $\hat{r}_i^L$ is calculated based on (\ref{eq:naive_basis}). By taking log likelihood of (\ref{eq:likelihood}) instead, we obtain:
\begin{equation}
L(\mathbf{w}) = - \dfrac{1}{2} \log ({2 \pi}) - \dfrac{1}{2}\log ({s_i}) - \dfrac{ \parallel y_i - \hat{r}_i^L \parallel^2}{2 s_i}  \enspace .
\end{equation}
The first term is a constant and is ignored, and the cost function is defined as minimizing the negative log likelihood:
\begin{equation}\label{eq:loss1}
C(\mathbf{w}) = \dfrac{1}{N} \sum_{i=1}^{N} \underbrace{\dfrac{ \parallel y_i - \hat{r}_i^L \parallel^2}{s_i}}_{\text{Data term}} + \underbrace{\log ({s_i}) \enspace}_{\text{Regl. term}} , \\
\end{equation}
\noindent where the scalar ${1/2}$ has been discarded. This cost function shares similarity with the aleatoric uncertainty loss proposed in \cite{kendall2017uncertainties}. The difference is that $s_i$ in our case depicts an uncertainty measure that encodes observation noise variance and the output confidence from NCNN, while in \cite{kendall2017uncertainties}, it is the variance of the noise. Note that this cost function can be derived using any error model from the exponential family, \eg Laplace distribution as in \cite{ilg2018uncertainty}. Next, we show the architecture design that is used for training our proposed probabilistic approach.


\subsection{Probabilistic NCNN Architecture}
Given a dataset that contains undisturbed data $\mathcal{Y}$ as groundtruth and a disturbed version $\dot{\mathcal{Y}}$ as input, we aim to train a network that produces the clean data given the disturbed one. An illustration for our full pipeline is shown in Figure \ref{fig:arch}. The first component estimates the input confidence from the disturbed input and feed both of them to the NCNN network. The output confidence from the last NCNN layer is fed to another compact UNet to estimate the noise parameter $\sigma^2_i$ and to produce $s_i$ in (\ref{eq:loss1}). Finally, the prediction from the NCNN network and the estimated uncertainty $s_i$ are fed to the loss.

Note that the noise variance estimation network takes only the output confidence from the NCNN as input, contrarily to existing approaches that estimate the uncertainty from the final prediction \cite{gustafsson2019evaluating,ilg2018uncertainty}. This indicates that our confidences can efficiently encode the uncertainty information, which is also demonstrated in the experiments section.



%% file: sec/exp.tex
To demonstrate the capabilities of our proposed approach, we evaluate it on the KITTI-Depth dataset \cite{uhrig2017sparsity} for the task of \emph{unguided} depth completion (no RGB guidance is used). We first compare against Bayesian Deep Learning approaches, \eg MC-Dropout \cite{gal2016dropout} and ensembling \cite{lakshminarayanan2017simple}, in terms of prediction accuracy and the quality of the uncertainty measure. Then, we show comparison against the conventional non-statistical approaches. Afterwards, we perform an ablation study for different components of our pipeline and we experiment with an ensemble of our proposed network. Finally, we demonstrate the generalization capabilities of our approach by evaluating it on multi-path interference correction \cite{guo2018tackling} and optical flow rectification. The source code is available on Github \footnote{\url{https://github.com/abdo-eldesokey/pncnn}}.


\subsection{Experimental Setup}
Our pipeline is illustrated in Figure \ref{fig:arch} and more details are given in the supplementary materials. We evaluate three variations of our network: our network where only the input confidence estimation part that is trained using the L1 or the L2 norm (\emph{NCNN-Conf}), our full network trained with the proposed loss in (\ref{eq:loss1}) (\emph{pNCNN}), and our full network trained with a modified version of the loss in (\ref{eq:loss1}), where we apply an exponential function to $s_i$ in the data term (\emph{pNCNN-Exp}). This modification is to robustify our loss to outliers violating the presumed Gaussian error model for the data term. Training was performed using the Adam optimizer with an initial learning rate of $0.01$ that is decayed with a factor of $10^{-1}$ every 3 epochs.

\begin{figure}[!t]
	\vspace*{-5mm}
	\includegraphics[width=\columnwidth]{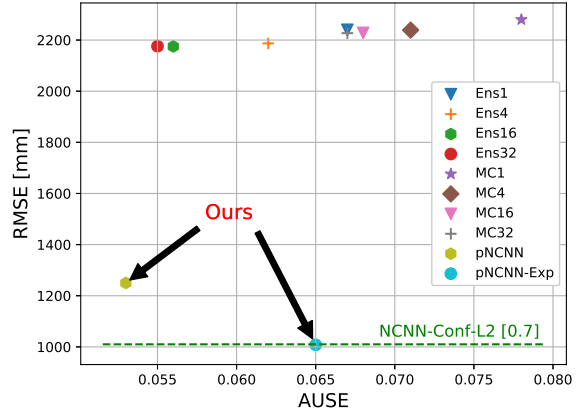}
	\caption{A comparison between statistical approaches in terms of RMSE and AUSE metrics where bottom-left is better. The two variations of our approach outperforms other methods w.r.t. RMSE and \emph{pNCNN} trained with (\ref{eq:loss1}) produces the best uncertainty measure. Note that \emph{NCNN-Conf-L2} only achieves AUSE of 0.7.}
	\label{fig:rmse_ause}
\end{figure}

\noindent \textbf{Evaluation Metrics} We use the following two measures:

\noindent \emph{Prediction Error} We use the error metrics from the KITTI-Depth \cite{uhrig2017sparsity} such as Mean Average Error (MAE), Root Mean Square Error (RMSE) and their inverses.	

\noindent \emph{Quality of Uncertainty} We use the sparsitification error plots and the area under sparsification error plots (AUSE) \cite{ilg2018uncertainty} as a measure for the quality of the uncertainty.


\subsection{Results Compared to Statistical Methods}

Gustafsson \etal \cite{gustafsson2019evaluating} evaluated the MC-Dropout \cite{gal2016dropout} and ensembling \cite{lakshminarayanan2017simple} by modifying the head of the Sparse-to-Dense (S2D) \cite{ma2019self} network to output the parameters of a Gaussian distribution. They evaluated an ensemble of 1-32 instances of S2D with 26M parameters each an taking the mean of these instances for the final prediction. Note that their network utilizes both depth and RGB images, while our approach consist of a \emph{single} network that is fully unguided and uses only depth data.

Figure \ref{fig:rmse_ause} shows a two-metric comparison with respect to AUSE and RMSE. Our \emph{NCNN-Conf} performs best in terms of RMSE, while it performs worst in terms of AUSE. On the other hand, our full network trained with the proposed loss, \emph{pNCNN}, produces the best uncertainty measure with an AUSE of \textbf{0.053} outperforming an ensemble of 32 networks. Moreover, it achieves a significantly lower RMSE than MC-Dropout and ensembling. However, it performs inferior to \emph{NCNN-Conf} in terms of RMSE with a moderate gap. The variation of our network that is trained with a modified loss, \emph{pNCNN-Exp}, closes this gap and performs on-par with \emph{NCNN-Conf} in terms of RMSE with a minor degradation of AUSE compared to \emph{pNCNN}.



\subsection{Results Compared to Non-Statistical Methods}
We also compare our proposed approach against the non-statistical unguided approaches. Table \ref{tab:kitti} summarizes the results on the test set of the KITTI-Depth dataset. Our \emph{NCNN-Conf-L1} outperforms all other methods on three out of four metrics when compared individually, except for \emph{Spade}, where we are better on two metrics and on-par on one metric. Note the improvement of our approach over the standalone \emph{NCNN}, where we achieve a performance boost of $\sim 45\%$ by providing more accurate input confidences. Our probabilistic model trained using a Gaussian error model  and a Laplace error model, \emph{pNCNN-Exp} trained with the modified loss performs equally good to the \emph{NCNN-Conf-L2}, but additionally providing proper output uncertainties.

\renewcommand{\arraystretch}{1.15}
\begin{table}[t]
	\small \selectfont
	\begin{center}
		\begin{tabular}{L{2.3cm} | C{0.8cm} C{0.8cm} C{0.8cm} C{0.6cm} C{0.5cm} }
			\specialrule{.2em}{.1em}{.1em} 
			& MAE [mm] & RMSE [mm] & iMAE [1/km] & iRMSE [1/km] & \#P \\
			\hline
			\hline
			SparseConv \cite{uhrig2017sparsity} & 481.27 & 1601.33 & 1.78 & 4.94 & 25k \\
			ADNN \cite{Chodosh2018} & 439.48 & 1325.37 & 3.19	& 59.39 & 1.7k	\\
			NCNN  \cite{bmvc} & 360.28 & 1268.22 & 1.52	& 4.67 & 0.5k \\
			S2D  \cite{ma2019self} & 288.64 & 954.36 & 1.35 & 3.21 & 26M	\\
			HMS-Net \cite{hms} & 258.48	& \textit{937.48} & 1.14& 2.93 & - \\
			SDC \cite{wvangansbeke_depth_2019} & 249.11 & \textbf{922.93} & 1.07 & 2.80 & 2.5M \\
			Spade  \cite{valeo} & \textit{248.32} & 1035.29 & \textbf{0.98} & \textbf{2.60} & 5.3M \\
			\hline
			\textbf{NCNN-Conf-L1 } & \textbf{228.53} & 988.57 & \textit{1.00} & \textit{2.71} & 330k \\
			\textbf{NCNN-Conf-L2} & 258.68 & 954.34 & 1.17 & 3.40 & 330k \\
			\textbf{pNCNN-Exp} & 251.77 & 960.05 & 1.05 & 3.37 & 670k \\
			\specialrule{.2em}{.1em}{.1em} 
		\end{tabular}
	\end{center}
	\caption{Quantitative results on the \emph{test} set of the KITTI-Depth for \emph{unguided} depth completion. \#P is the number of parameters.}
	\label{tab:kitti}
\end{table}

\subsection{Ablation Study}

\begin{figure}[!t]
	\includegraphics[width=\columnwidth]{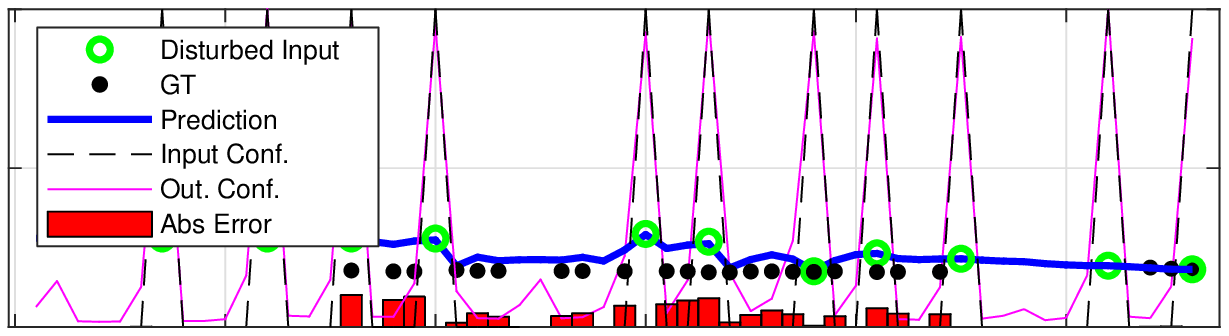} %
	\includegraphics[width=\columnwidth]{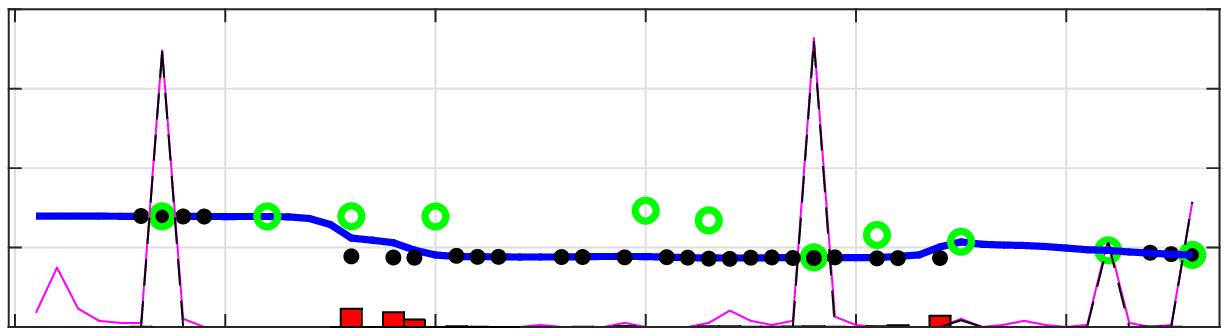} %
	\includegraphics[width=\columnwidth]{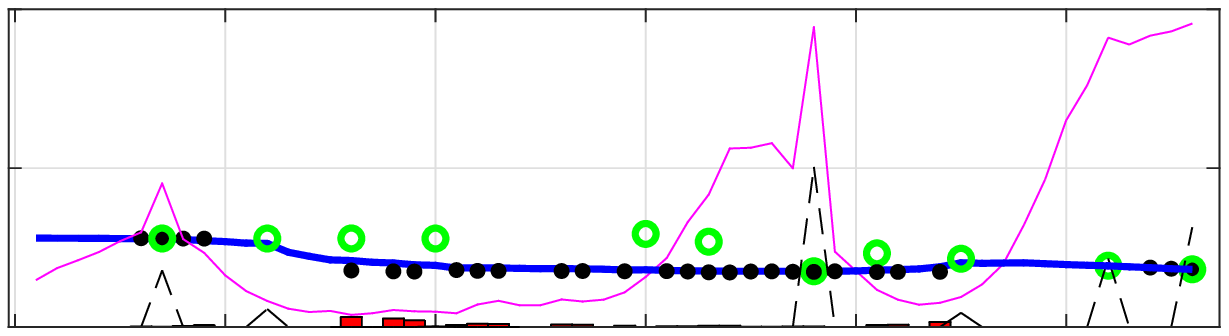} %
	\caption{A qualitative example from the KITTI-Depth dataset showing the impact of each component of our proposed approach. First row is the standard \emph{NCNN}, the second is \emph{NCNN-Conf-L2}, and the third is \emph{pNCNN}.}
	\label{fig:abl}
\end{figure}

First we show the impact of each component of our proposed network on a qualitative example from the KITTI-Depth dataset. Figure \ref{fig:abl} shows an example where the input measurements do not coincide with the groundtruth. The standard NCNN assigns 1-confidences to all measurements, which results in a corrupted prediction (first row). When we apply our input confidence estimation, the disturbed measurements are successfully identified and assigned zero  confidence (second row). However, the output confidence is almost identical to the input confidence and shows no strong correlation with the accuracy. When we apply our full pipeline, the disturbed measurements are identified and the output uncertainty becomes highly correlated with the prediction error (third row).

Next, we show in Table \ref{tab:abl} the impact of modifying different components of our pipeline. When the confidence estimation is discarded in \emph{w/o conf-est} and binary input confidence is used, the RMSE is degraded, while the network still manages to achieve good AUSE. Similarly, when the noise variance estimation network is discarded in \emph{w/o var-est}, the RMSE is severely degraded as the input confidence estimation network tries to make up for the absence of the variance estimation network. When the final prediction from the NCNN is fed along with the output confidence to the noise variance estimation network in \emph{w depth-pred}, no improvement is gained in terms of AUSE. This demonstrates that our uncertainty measure efficiently encode the uncertainty information in the NCNN confidence stream without looking at the prediction. Finally, when we employ a Laplace error model for the loss in \emph{w Laplace-loss}, \ie, the L1 norm for residuals, the MAE improves, while AUSE is degraded since it is calculated based on the RMSE.

\renewcommand{\arraystretch}{1.1}
\begin{table}[t]
	\begin{center}
		\begin{tabular}{L{2.5cm} | C{1.5cm}  C{1.5cm}  C{1.0cm}}
			\specialrule{.2em}{.1em}{.1em} 
			 & RMSE 	& MAE  	& AUSE 	\\ 
			\hline
			\hline
			pNCNN & 1237.65 & 283.41 & 0.055 \\ 
			  \ - w/o conf-est & 1540.00 & 405.00 & 0.058 \\ 
			  \ - w/o var-est & 1703.50  & 604.10 & 0.123 \\ 
			  \ - w depth-pred & 1215.64 & 292.68 & 0.055   \\ 
			  \ - w Laplace-loss & 1272.32 & 248.26  & 0.089  \\ 
			\specialrule{.2em}{.1em}{.1em} 
		\end{tabular}
	\end{center}
	\caption{The results for the ablation study when trained on a subset of the training set evaluated on the selected validation set of the KITTI-Depth dataset.} 
	\label{tab:abl}
\end{table}


\subsection{Ensemble of pNCNN}
To examine whether our probabilistic approach can be extended to a fully Bayesian approach, we form an ensemble of four \emph{pNCNN} network that were initialized randomly and trained on random subset of the KITTI-Depth dataset. We evaluate multiple fusion approaches which are summarized in Table \ref{tab:ens}. Fusion by selecting the most confident pixel from each network, \emph{maxConf}, achieves the best results, outperforming taking the mean, which is commonly used. Taking a weighted mean using confidences, \emph{wMean}, or a maximum likelihood estimation, \emph{MLE}, also gives better results than the standard mean. This demonstrated the potential of using the proposed output confidences in more sophisticated fusion schemes.

\renewcommand{\arraystretch}{1.1}
\begin{table}
	\begin{center}
		\begin{tabular}{L{0.9cm} | C{0.9cm}  C{0.8cm} !{\vrule width 1.5pt}  L{1.3cm} | C{0.9cm} C{0.8cm}}
			\specialrule{.2em}{.1em}{.1em} 
			& RMSE 	& MAE  & \textbf{Fusion} & RMSE & MAE	\\ 
			\hline
			\hline
			Net-1 & 1337.5 & 290.5 & Mean & 1287.3 & 290.5 \\
			Net-2 & 1325.1 & 303.1 & wMean & 1261.3 & 285.9 \\
			Net-3 & 1315.1 & 296.9 & maxConf & \textbf{1260.7} & {283.8} \\
			Net-4 & 1321.1 & 288.3 & MLE & 1264.1 & \textbf{282.4} \\
			\specialrule{.2em}{.1em}{.1em} 
		\end{tabular}
	\end{center}
	\caption{Fusion schemes for an ensemble of \emph{pNCNN} trained on a subset of the KITTI-Depth and evaluated on the selected validation set. \emph{MLE} refers to Maximum Likelihood Estimation.} 
	\label{tab:ens}
\end{table}


\subsection{Mutli-Path Interference (MPI) Correction}

To demonstrate the generalization capabilities of our approach on other kinds of noise, we evaluate it on depth data from a Time-of-Flight (ToF) camera, \ie Kinect2, that suffers from MPI. We use the FLAT dataset \cite{guo2018tackling} for this purpose which provides raw measurements for three different frequencies and phases. We use the libfreenect2 \cite{libfreenect} to calculate the depth from the measurements and we compare against applying the bilateral filtering on the noisy depth.

Table \ref{tab:flat} summarizes the results, where we outperform the Bilateral filtering with a significant margin in terms of RMSE error when evaluated both on noisy and clean data with no MPI. Bilateral filtering on the other hand performs worse than doing no processing as it assigns zeros to pixels close to edges. When edges are not considered for evaluation, bilateral filtering improves the results slightly, but is outperformed by our approach.

\renewcommand{\arraystretch}{1.1}
\begin{table}[b]
	\small \selectfont
	\begin{center}
		\begin{tabular}{l | c | c | c }
			\specialrule{.2em}{.1em}{.1em} 
		 RMSE [mm]&  Ours & Biateral & No-Proc \\
			\hline
			\hline
			No-MPI    &  \textbf{231} &   444 &  415  \\ 
			\hline
			MPI     &  \textbf{283}  &  429 &  449	 \\ 
			\hline
			No-MPI-Masked     &  \textbf{175} &  263 &  288 \\ 
			\hline
			MPI-Masked    &  \textbf{205} &  282 &  299 \\ 
			\specialrule{.2em}{.1em}{.1em} 
		\end{tabular}
	\end{center}
	\caption{The RMSE error in millimeters for Multi-Path Interference (MPI) correction on the FLAT dataset \cite{guo2018tackling}. \emph{No-Proc} refers to evaluating the depth without any processing. The masked version disregards edges from the evaluation. }
	\label{tab:flat}
\end{table}


\subsection{Sparse Optical Flow Rectification}
\begin{figure}[t]
	\begin{tabular}{C{0.5\columnwidth} C{0.5\columnwidth}}
		{\includegraphics[width=0.45\columnwidth]{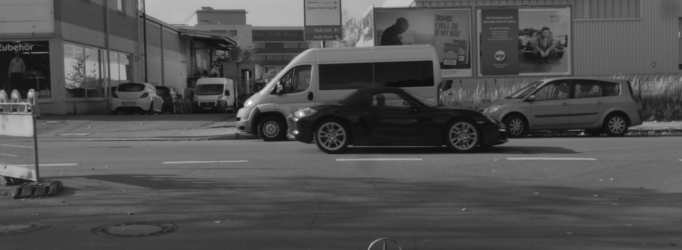}} &
		\fbox{\includegraphics[width=0.45\columnwidth]{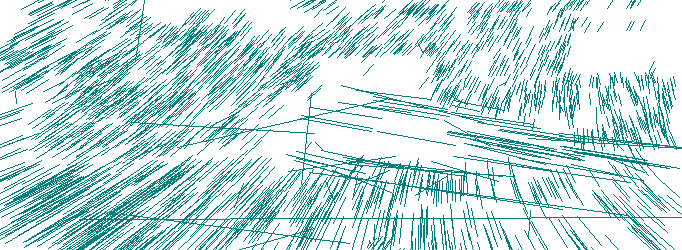}} \\
		\fbox{\includegraphics[width=0.45\columnwidth]{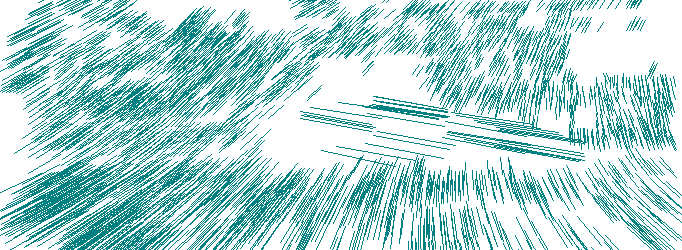}} &
		\fbox{\includegraphics[width=0.45\columnwidth]{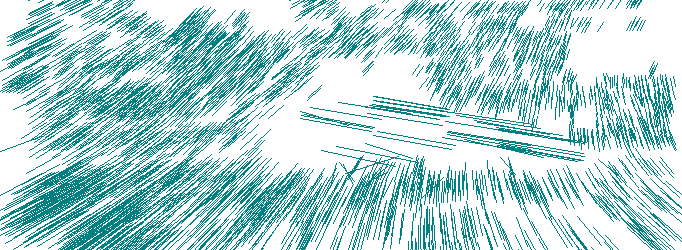}}	\\
		\vspace{1mm}
	\end{tabular}
	\caption{
		Qualitative example for optical flow outliers rejection. In right-bottom order, RGB frame, raw flow input, groundtruth flow, and estimated flow. 
	}
	\label{fig:opt}
\end{figure}

We generate the input flow by applying the Lucas-Kanade method \cite{lucas1981iterative} to pairs of images from driving sequences. The groundtruth is produced by geometrical verification over several frames under a multiple rigid body assumption \cite{persson2015robust}. Figure \ref{fig:opt} shows an example for rectifying the corrupted measurement and densifying the flow field. More results are given in the supplementary materials.
\renewcommand{\arraystretch}{0.5} 
\setlength{\tabcolsep}{0.1pt} 

\subsection{What happens if the input is undisturbed?}
\begin{figure}[t]
	\begin{minipage}[b]{0.51\linewidth}
		\centering
		\includegraphics[width=0.9\textwidth]{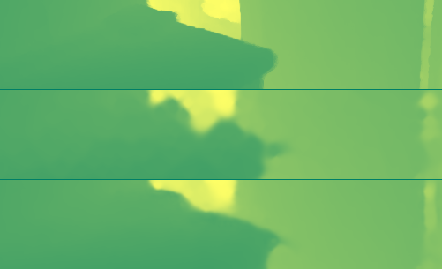}%
		\caption{A qualitative example from the NYU dataset \cite{Silberman12}. Top-to-bottom: groundtruth, NCNN \cite{bmvc}, NCNN-Conf.}
		\label{fig:nyu}
	\end{minipage}
	\hspace{0.05cm}
	\begin{minipage}[b]{0.45\linewidth}
		\begin{tabular}{L{1.7cm} | C{0.9cm} | C{0.7cm} }
			\specialrule{.2em}{.1em}{.1em} 
			 & {\small \rotatebox[origin=c]{90}{  RMSE  } } & \small \rotatebox[origin=c]{90}{ MAE  }   \\
			\hline
			\small NCNN \cite{bmvc}  & 0.165 & 0.07  \\
			\hline
			\small NCNN-Conf  & \textbf{0.135} & \textbf{0.05}  \\
			\hline
			\small pNCNN & 0.144 & 0.06  \\
			\specialrule{.2em}{.1em}{.4em} 
			
		\end{tabular}
		\caption{Quantitative results on the NYU dataset \cite{Silberman12} in meters.}
		\label{tab:nyu}
	\end{minipage}
\end{figure}

\vspace{-2mm}
An essential question is how our confidence estimation network will perform if the input data is not disturbed? To answer this question, we train our network \emph{NCNN-Conf} and \emph{pNCNN} on the NYU dataset \cite{Silberman12}, where the input is sampled from the groundtruth depth. We use 1000 depth points sampled uniformly with a sparsity level of {$0.6\%$}. Figure \ref{fig:nyu} and Table \ref{tab:nyu} show that both our methods surprisingly improves the results compared to the standalone NCNN \cite{bmvc}. This is a result of allowing the confidence estimation network to assign proper confidences to points based on their proximity to edges similar to non-linear filtering. This leads to sharper edges and better reconstruction of objects.
\renewcommand{\arraystretch}{1.1} 
\setlength{\tabcolsep}{2pt} 

%% file: supplementary.tex
\begin{center}
\LARGE{Supplementary Material for \linebreak Uncertainty-Aware CNNs for Depth Completion: \linebreak Uncertainty from Beginning to End}
\end{center}

\section{Implementation Details}
\input{sec/impl.tex}

\section{Ensemble methods}

\input{sec/ensemble.tex}

\section{Additional Results}
\input{sec/exp_supp.tex}

%% file: sec/impl.tex
In this section, we give more details on the implementation of our proposed method such as the loss function and the design of the confidence estimation and the noise variance estimation networks. 


\subsection{The Loss Function}
We drove a loss function to train the proposed probabilistic normalized convolutional neural networks \emph{(pNCNN)}, which reads:
\begin{equation}\label{eq:loss1}
C(\mathbf{w}) = \dfrac{1}{N} \sum_{i=1}^{N} \underbrace{\dfrac{ \parallel y_i - \hat{r}_i^L \parallel^2}{s_i}}_{\text{Data term}} + \underbrace{\log ({s_i}) \enspace}_{\text{Regl. term}} , \\
\end{equation}

\noindent where $s_i$ is the proposed uncertainty measure and it is equal to $\sigma_i^2 / \langle \mathbf{a} | \mathbf{c} \rangle$. For convenience and numerical stability, we modify the regularization term so that $s_i$ becomes consistent with the data term. This leads to:
\begin{equation}\label{eq:loss2}
C(\mathbf{w}) = \dfrac{1}{N} \sum_{i=1}^{N}{\dfrac{ \parallel y_i - \hat{r}_i^L \parallel^2}{s_i}} - {\log (\dfrac{1}{s_i}) \enspace} , \\
\end{equation}

\noindent This can be expanded using the definition of $s_i$:
\begin{equation}\label{eq:loss3}
C(\mathbf{w}) = \dfrac{1}{N} \sum_{i=1}^{N}  \dfrac{\langle \mathbf{a}^L | \mathbf{c}^L \rangle }{\sigma_i^2 } \parallel y_i - \hat{r}_i^L \parallel^2 - \log (\dfrac{\langle \mathbf{a}^L | \mathbf{c}^L \rangle }{\sigma_i^2 }) \enspace 
\end{equation}
\noindent where $\mathbf{a}^L, \mathbf{c}^L$ are the learned applicability and the output confidence of the last normalized convolution layer $L$ respectively. This expansion makes it clear that our proposed uncertainty measure depends both on the output confidence from the normalized convolution layer and observations noise variance. A higher noise variance will reduce the output confidence from the NCNN and vice versa. This indicates that our proposed uncertainty measure encodes the single observation noise as well as the confidence with respect to the neighboring pixels.


\subsection{The Architecture}
\begin{figure}[t]	
	\centering
    {\includegraphics[width=0.55\textwidth]{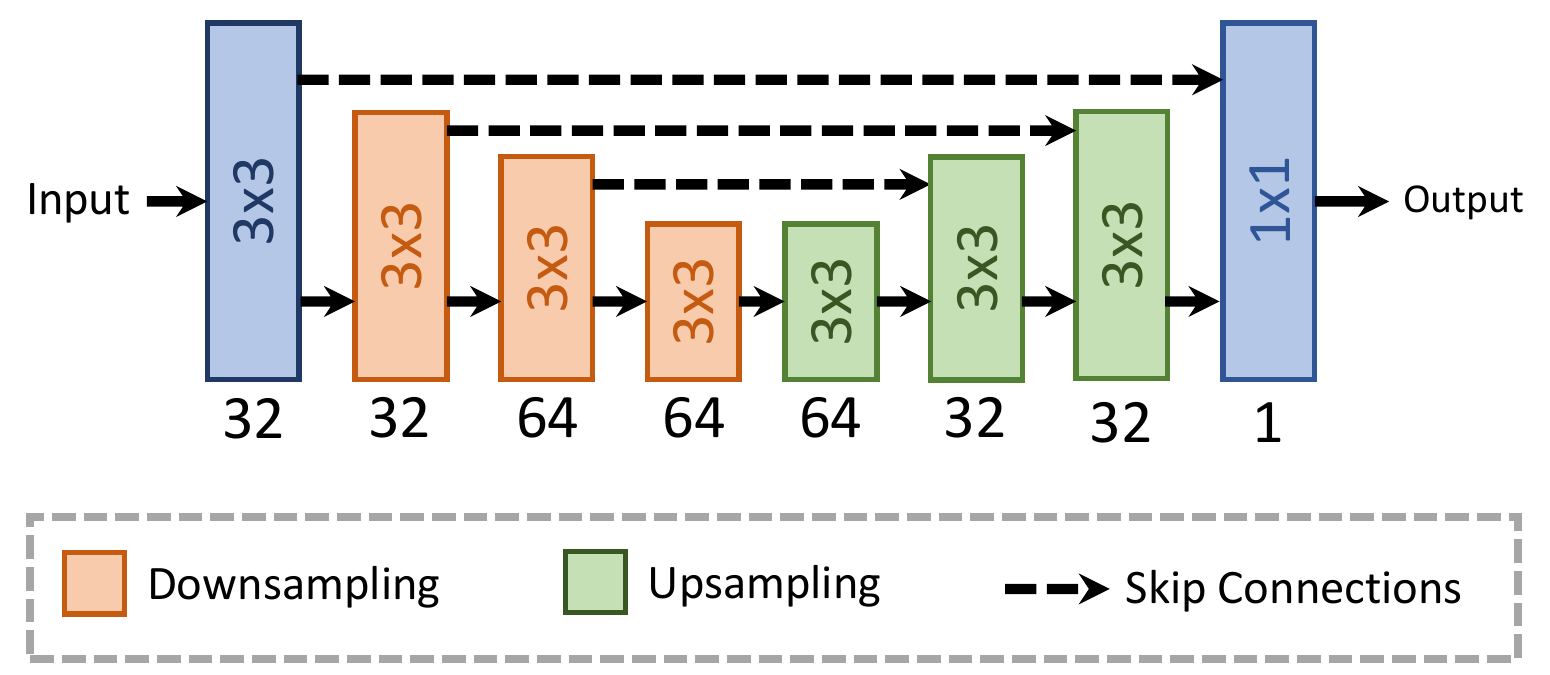}}
     \vspace{0.2cm}    
	{\includegraphics[width=0.34\textwidth]{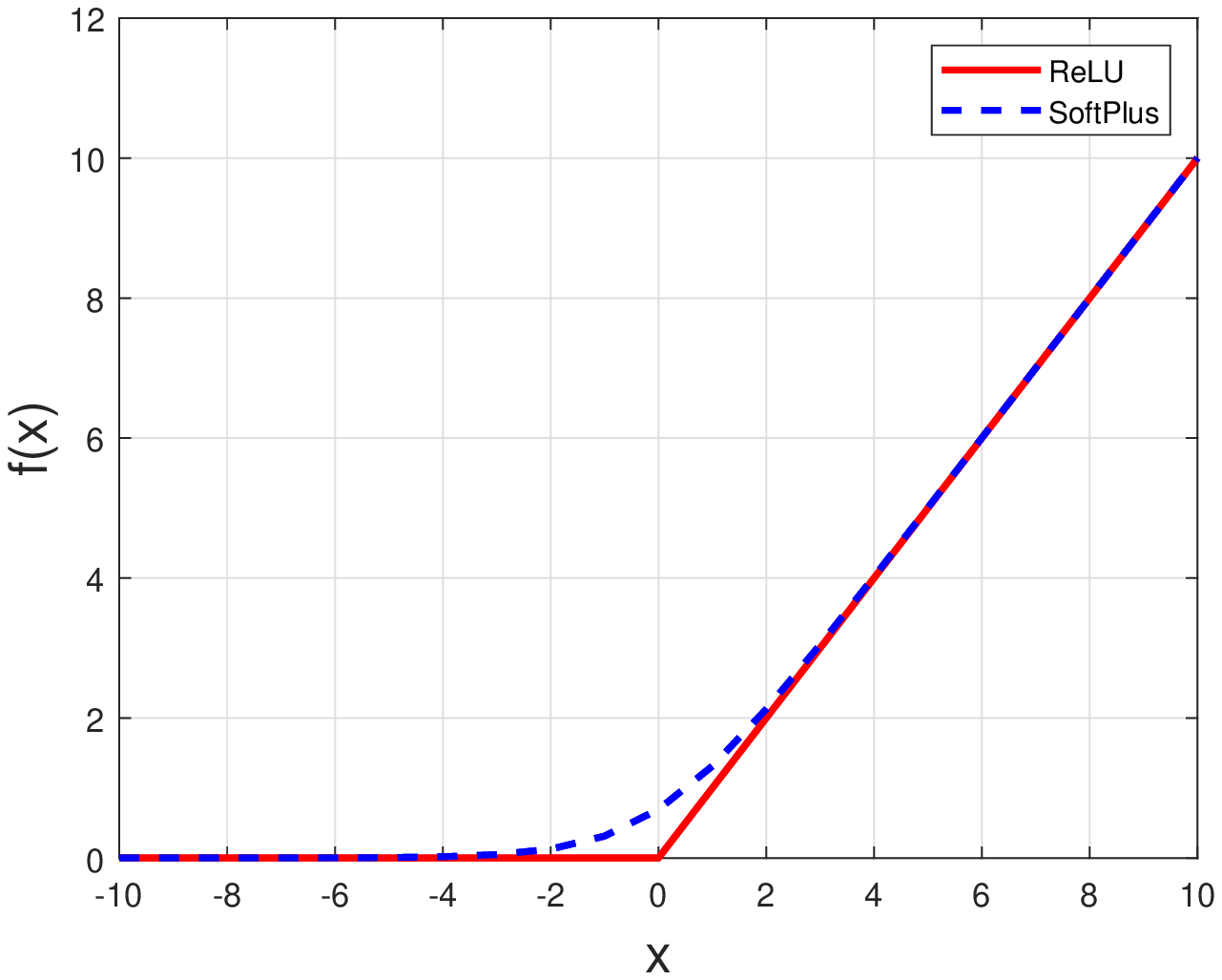}}
	\caption{(a) The proposed compact UNet used for the confidence estimation network and the noise variance estimation network. (b) The SoftPlus activation used at the final layer in comparison with the ReLU activation.}
	\label{fig:unet}
\end{figure}

We propose to learn the input confidence using a compact UNet \cite{unet} that is trained end-to-end with a normalized convolutional neural network (NCNN) \cite{pami}. We also learn observations noise variance using a similar UNet. The design of this UNet is shown in Figure \ref{fig:unet} and it is identical for both networks. It is worth mentioning that this network has only 3 scales compared to original UNet which has 4 scale, since we found empirically that the 4th scale does not improve the estimation. The number of channels per convolution layer was significantly reduced for computational efficiency.

The choice of the activation for the last layer is crucial since it must produce valid range of values for confidences $\left[0, \infty \right[$. We choose the SoftPlus function (Shown in Figure \ref{fig:unet}) due to its similarity to the ReLU activation. However, it does not suffer from the gradient discontinuity at zeros.

%% file: sec/ensemble.tex
In the main paper, we evaluate different fusion schemes for an ensemble of our network \emph{pNCNN}.  We showed that all fusion schemes utilizing our proposed uncertainty measure outperform the commonly used fusion using the standard mean. Here, we give the definition for the evaluated fusion schemes.

\subsection{The Standard Mean}
\noindent The Mean fusion method refers to the average over the predictions $y_i^k$ at pixel $i$:
\begin{equation}
\hat{y}_i = \frac{1}{N} \sum_{k=1}^{N} y_i^k \enspace .
\end{equation}

\subsection{The Weighted Mean}
\noindent Since the mean fusion does not take into account the uncertainties, we weight the predictions using their confidences $c_i^k$:
\begin{equation}
\hat{y}_i = \frac{1}{\sum_{k=1}^{N} c_i^k} \sum_{k=1}^{N} c_i^k y_i^k \enspace .
\end{equation}

\subsection{Max Voting}
\noindent Another commonly used voting scheme is to select the most confident prediction $k_i = \arg_m \max c_i^m$

\begin{equation}
\hat{y}_i = y_i^{k_i} \enspace .
\end{equation}

\subsection{Maximum Likelihood Estimate}
\noindent We can interpret our predictors as components of a Gaussian Mixture Model. If the prediction corresponds to the mean and the confidence corresponds to the unnormalized mixture weights, we can write the likelihood of a prediction $\hat{x}$ given predictions $y^k$ from the networks as:
\begin{equation}
l(\hat{x}_i) = \frac{1}{\sum_{k=1}^{N} c_i^k} \sum_{k=1}^{N} \frac{c_i^k}{\sqrt{2\pi v^2}} \exp \left( \frac{\|\hat{x}_i - y_i^k\|^2}{2 v^2} \right) \enspace .
\end{equation}

\noindent We can formulate an inference procedure based on the MLE for each pixel $i$ as:

\begin{equation} 
\hat{y}_i = \operatorname*{arg\,max}_{\hat{x}_i} \sum_{k=1}^{N} \frac{c_i^k}{v_i} \exp \left(\frac{\parallel \hat{x}_i - y_i^k\parallel^2 }{2 v_i^2}\right) \enspace ,
\end{equation}

\noindent \textbf{Optimization Procedure} The likelihood function of a Gaussian Mixture Model is in general non-convex. However, for the 1D case, the number of modes is constrained to at most the number of components in the mixture \cite{carreira2003number}.
Since it is guaranteed that the global maxima will be found if all local maximas are explored, we optimize the objective starting from each of the predictions. We use the ADAM optimizer with a maximum amount of steps set to 500. And we select the maximum of the local maximas which were found. Note that since we do not explicitly estimate the variances of the components we set $v^2$ to 0.1 for our experiments. 



%% file: sec/exp_supp.tex
\begin{figure}[b]
	\centering
	\includegraphics[width=\textwidth]{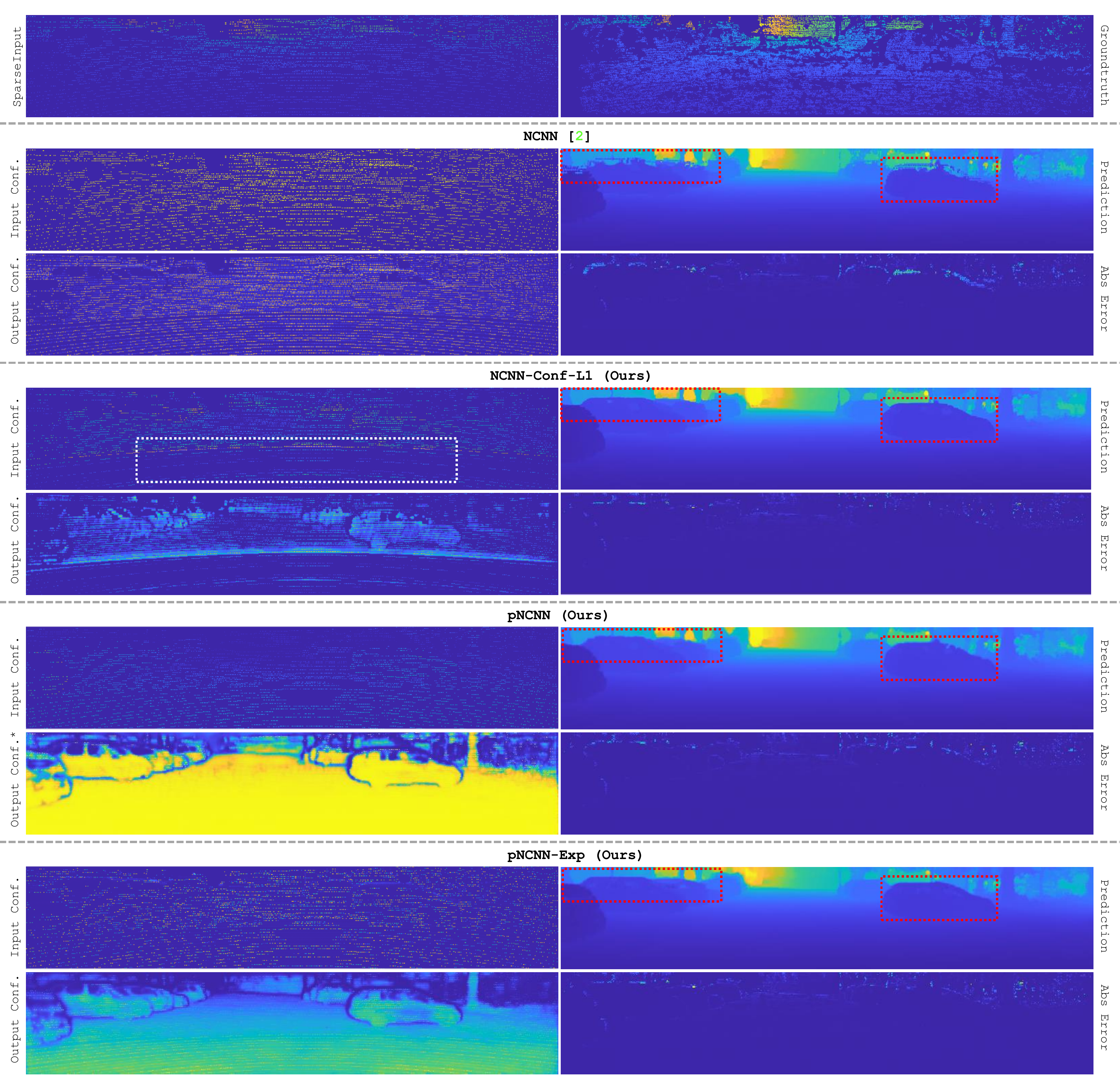}
	\caption{A qualitative example from the selected validation set of the KITTI-Depth dataset \cite{uhrig2017sparsity}. * denotes logarithmically scaled.}
	\label{fig:kitti_qual1}
\end{figure}

In this section, we show additional results for all the experiments in the paper. First, we show some qualitative examples on the KITTI-Depth dataset \cite{uhrig2017sparsity}. Then, we show the sparsificiation plots for our proposed uncertainty measure that were used to calculate the AUSE metric. Afterwards, we show some qualitative examples for multi-path interference correction and sparse optical flow rectification. Finally, we show illustrations on the NYU dataset \cite{Silberman12} for the case of undisturbed input data.

\subsection{Qualitative Results for The KITTI-Depth dataset}

\begin{figure}[t]
	\centering
	\includegraphics[width=\textwidth]{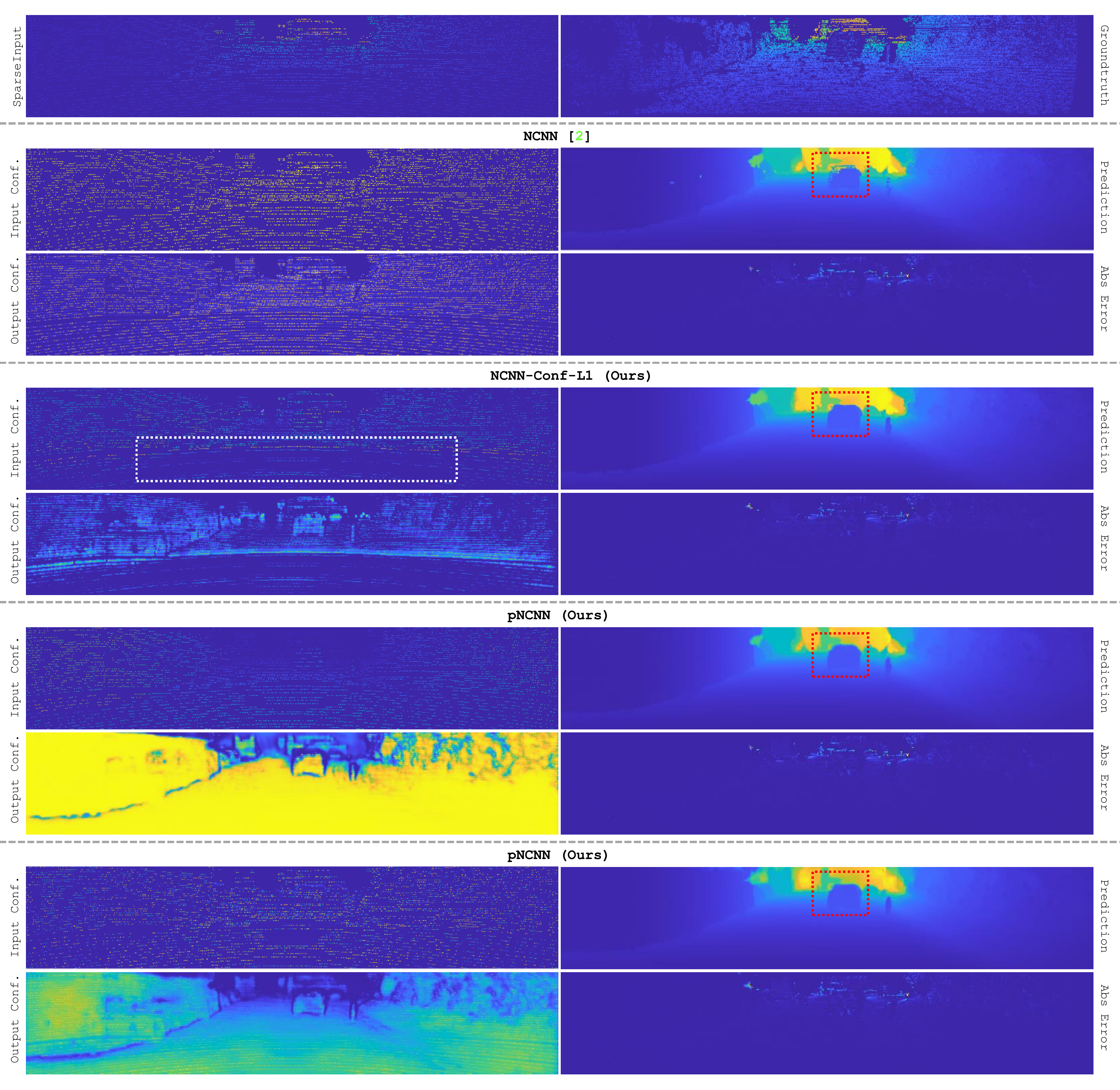}
	\caption{A second qualitative example from the selected validation set of the KITTI-Depth dataset \cite{uhrig2017sparsity}}
	\label{fig:kitti_qual2}
\end{figure}

Figure \ref{fig:kitti_qual1} and \ref{fig:kitti_qual2} show qualitative examples for \emph{NCNN} \cite{bmvc}, our proposed \emph{NCNN-Conf-L1}, \emph{pNCNN}, and \emph{pNCNN-Exp} from the selected validation set of the KITTI-Depth \cite{uhrig2017sparsity} dataset. \emph{NCNN} assigns binary confidence to the input, which results in artifacts at regions with disturbed measurements especially edges (indicated with red squares). Our proposed \emph{NCNN-Conf-L1} on the other hand, learns a proper input confidence which discards input measurements that causes the prediction error to increase. This causes the final prediction to be artifact-free and sharp along edges. It is worth mentioning that our input confidence estimation learned to discard some of the true measurements (indicated with the white squares) as well in order to produce smoother surfaces. Those discarded measurements are compensated for using other measurements on the end points of the same surface.

It is clear the output confidence from \emph{NCNN-Conf-L1} is a densified version of the estimated input confidence. But it does not provide full uncertainty information for all observations in the prediction. Our proposed \emph{pNCNN} addresses this problem and produces a reliable uncertainty measure for all observations. However, the prediction error at some disturbed measurements increase where the presumed Gaussian error model does not hold (indicated with the red squares if Figure \ref{fig:kitti_qual1}). By applying the exponential function to $s_i$ in the data term of the loss in \emph{pNCNN-Exp}, the network focuses more on minimizing the prediction error for those disturbed measurements and produces a better prediction. Note that  the range for the certainty measure changes with \emph{pNCNN-Exp} due to the exponential scaling.

\begin{figure}[t]
	\centering
	{\includegraphics[width=0.33\textwidth]{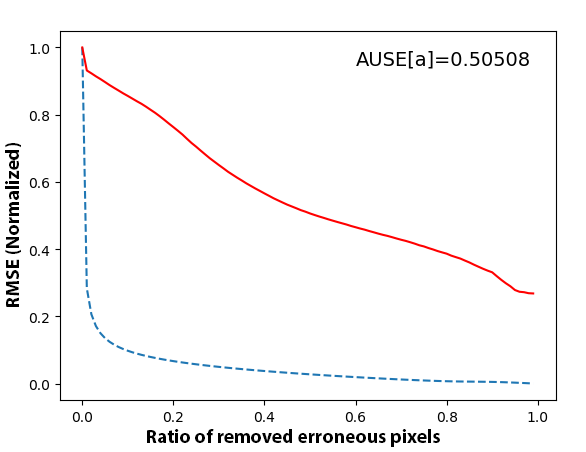}}
	{\includegraphics[width=0.33\textwidth]{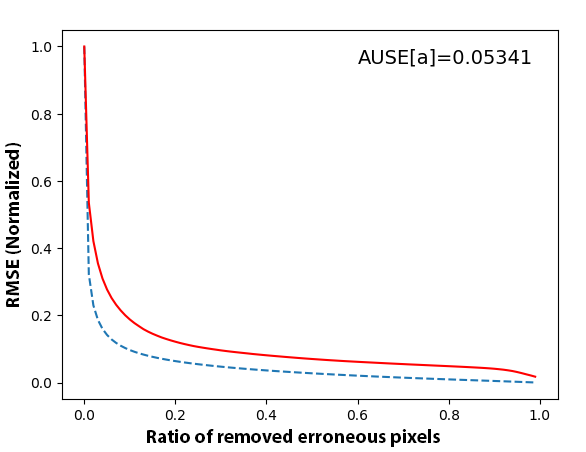}}
	{\includegraphics[width=0.33\textwidth]{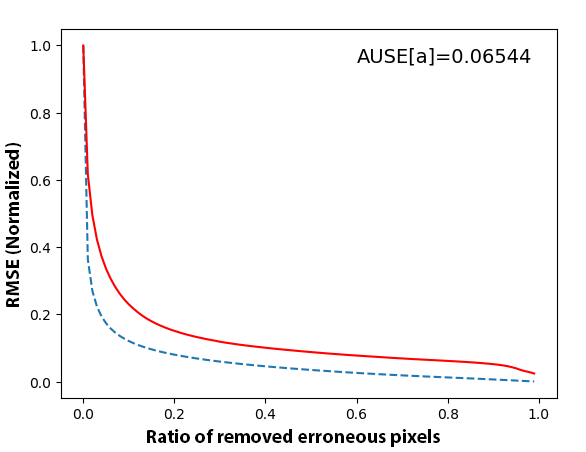}}
	\caption{Sparsification plots for (a) \emph{NCNN-Conf}, (b) \emph{pNCNN}, and (c) \emph{pNCNN-Exp}. The blue curve is the oracle and AUSE is the area between the two curves.}
	\label{fig:spep}
\end{figure}


\subsection{The Quality of the Proposed Uncertainty Measure}
To examine the quality of our proposed uncertainty measure, we look at the commonly used sparsification plots \cite{ilg2018uncertainty}. Sparsification plots show how efficiently the uncertainty measure discards the erroneous measurements. The baseline in this case is the prediction error itself, which is denoted as \emph{the oracle}. Sparsification plots for \emph{NCNN-Conf-L1},  \emph{pNCNN}, and \emph{pNCNN-Exp} are shown in Figure \ref{fig:spep}. The uncertainty measure from \emph{NCNN-Conf-L1} is not correlated with the oracle as the classical normalized convolution framework does not constitute any probabilistic properties. Our proposed probabilistic normalized convolution \emph{pNCNN} on the other hand, produces an accurate uncertainty measure that is very similar to the error oracle. The modified version \emph{pNCNN-Exp} also produces a high-quality uncertainty measure, but with a better handling of outliers.

\subsection{Multi-Path Interference Correction}

\begin{figure}[t]
	\centering
	\begin{tabular}{ C{0.25\columnwidth} C{0.25\columnwidth} C{0.25\columnwidth} C{0.25\columnwidth} }
		{ \footnotesize Groundtruth} & { \footnotesize Noisy depth} & 	{ \footnotesize pNCNN} & { \footnotesize Conf} \\
		\includegraphics[angle=-90,width=0.25\columnwidth]{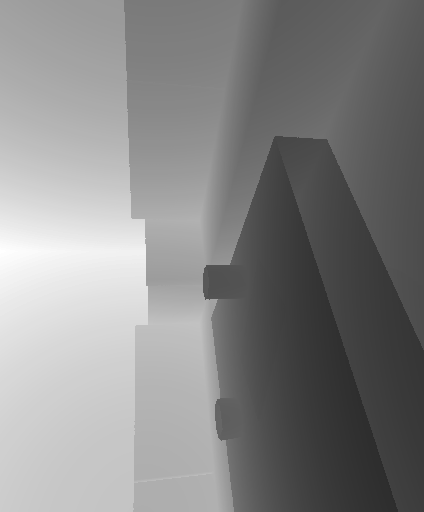} & 
		\includegraphics[angle=-90,width=0.25\columnwidth]{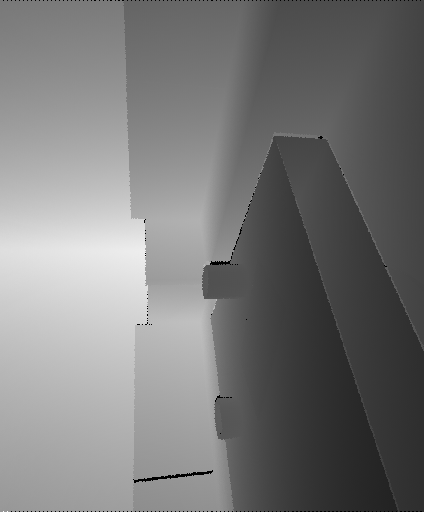} & 
		\includegraphics[angle=-90,width=0.25\columnwidth]{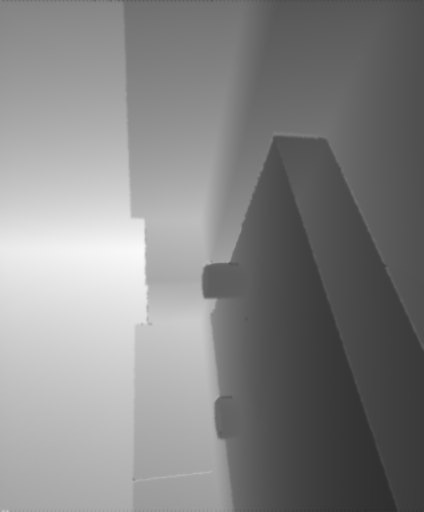} &
		\includegraphics[angle=-90,width=0.25\columnwidth]{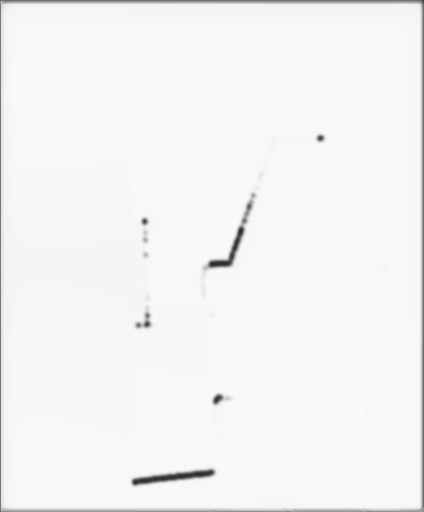}   \\
		\includegraphics[angle=-90,width=0.25\columnwidth]{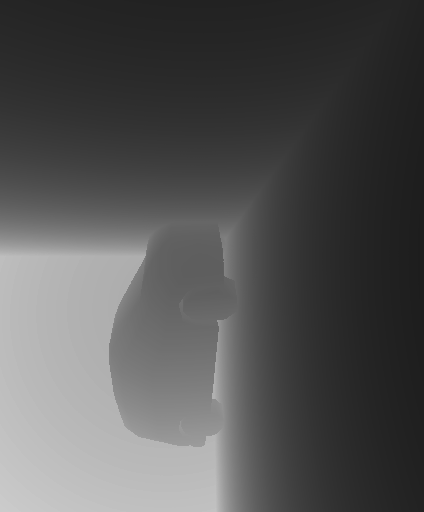} & 
		\includegraphics[angle=-90,width=0.25\columnwidth]{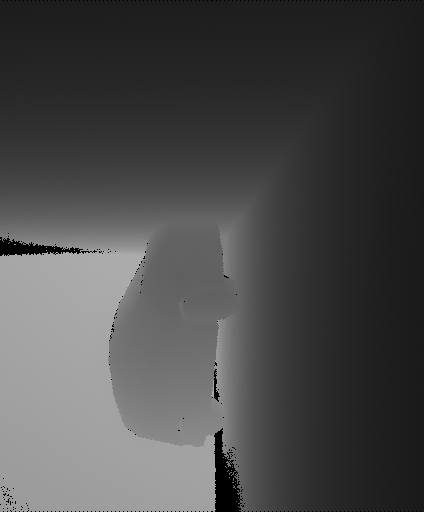} & 
		\includegraphics[angle=-90,width=0.25\columnwidth]{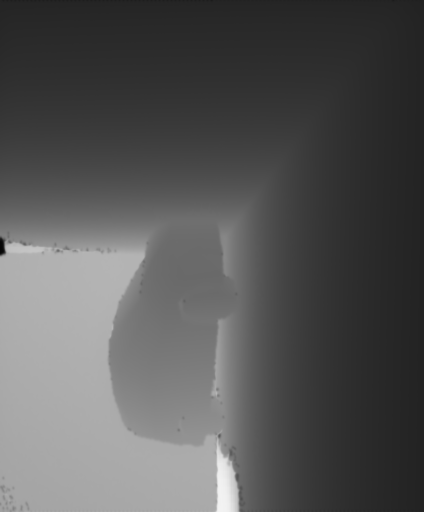} &
		\includegraphics[angle=-90,width=0.25\columnwidth]{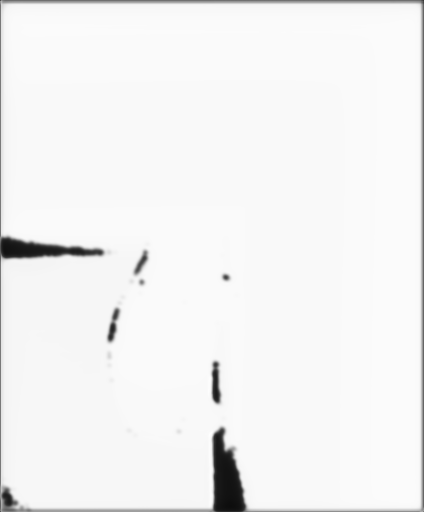}   \\
	\end{tabular}
	\caption{A qualitative example from the FLAT dataset showing the predicted output and the confidence  of the proposed approach.
	}
	\label{fig:flat}
\end{figure}

Figure \ref{fig:flat} shows two qualitative results for the FLAT dataset. The first row, shows a scene with small areas of missing data. These areas are well handled by the \emph{pNCNN} and the confidences clearly shows the uncertainty that exist in these areas and on edges.
The scene in the second row illustrates the effect of larger areas of missing data. These areas are missing too much data for the network to handle. As such, the output confidences is used to mask these parts of the signal. 
This illustrates the strength of our formulation in handling both smaller areas were the missing data can be extrapolated and larger areas where high uncertainty is assigned.

\renewcommand{\arraystretch}{0.2} 
\setlength{\tabcolsep}{0.5pt} 
\setlength{\tabcolsep}{5pt} 



\subsection{What happens when the input is undisturbed?}
\begin{figure}
	\centering
	\begin{tabular}{C{0.24\columnwidth} C{0.24\columnwidth} | C{0.24\columnwidth} C{0.24\columnwidth} } 
		{ \footnotesize NCNN-Conf-L1} & { \quad \footnotesize NCNN \cite{bmvc}} & { \footnotesize NCNN-Conf-L1} & { \quad \footnotesize NCNN \cite{bmvc}}  \\
		\includegraphics[width=0.24\columnwidth]{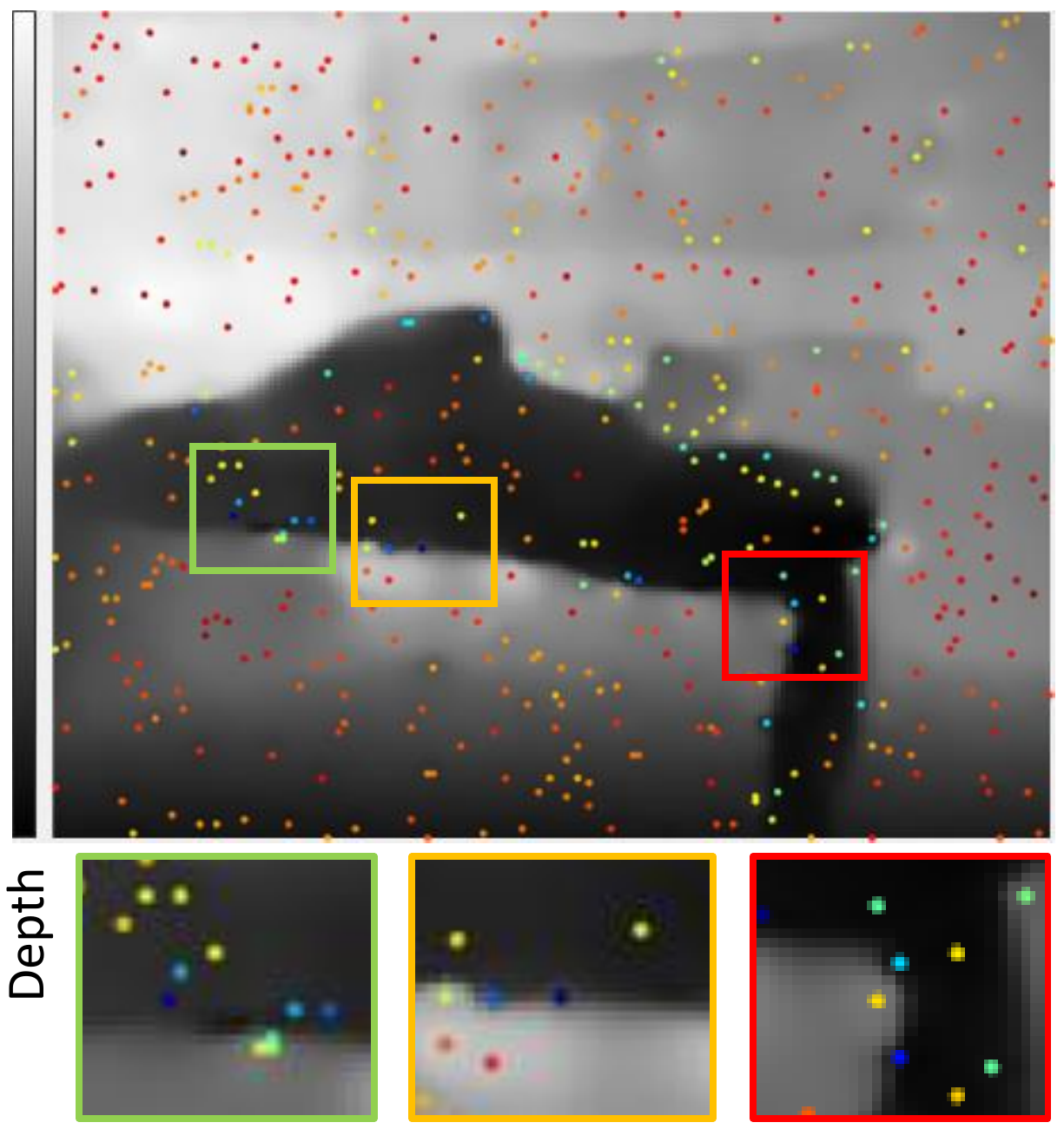} & 
		\includegraphics[width=0.24\columnwidth]{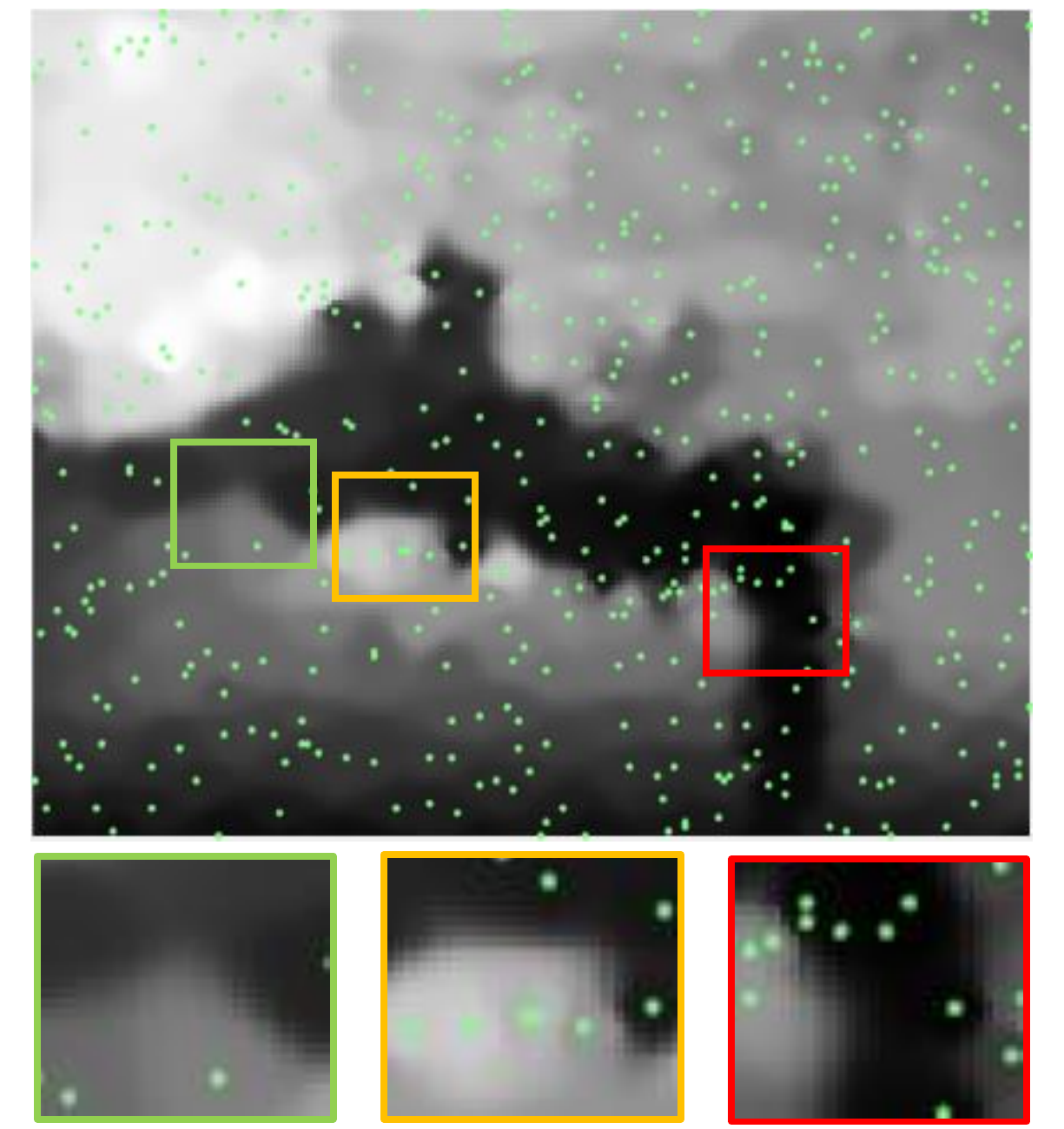} & 
		\includegraphics[width=0.24\columnwidth]{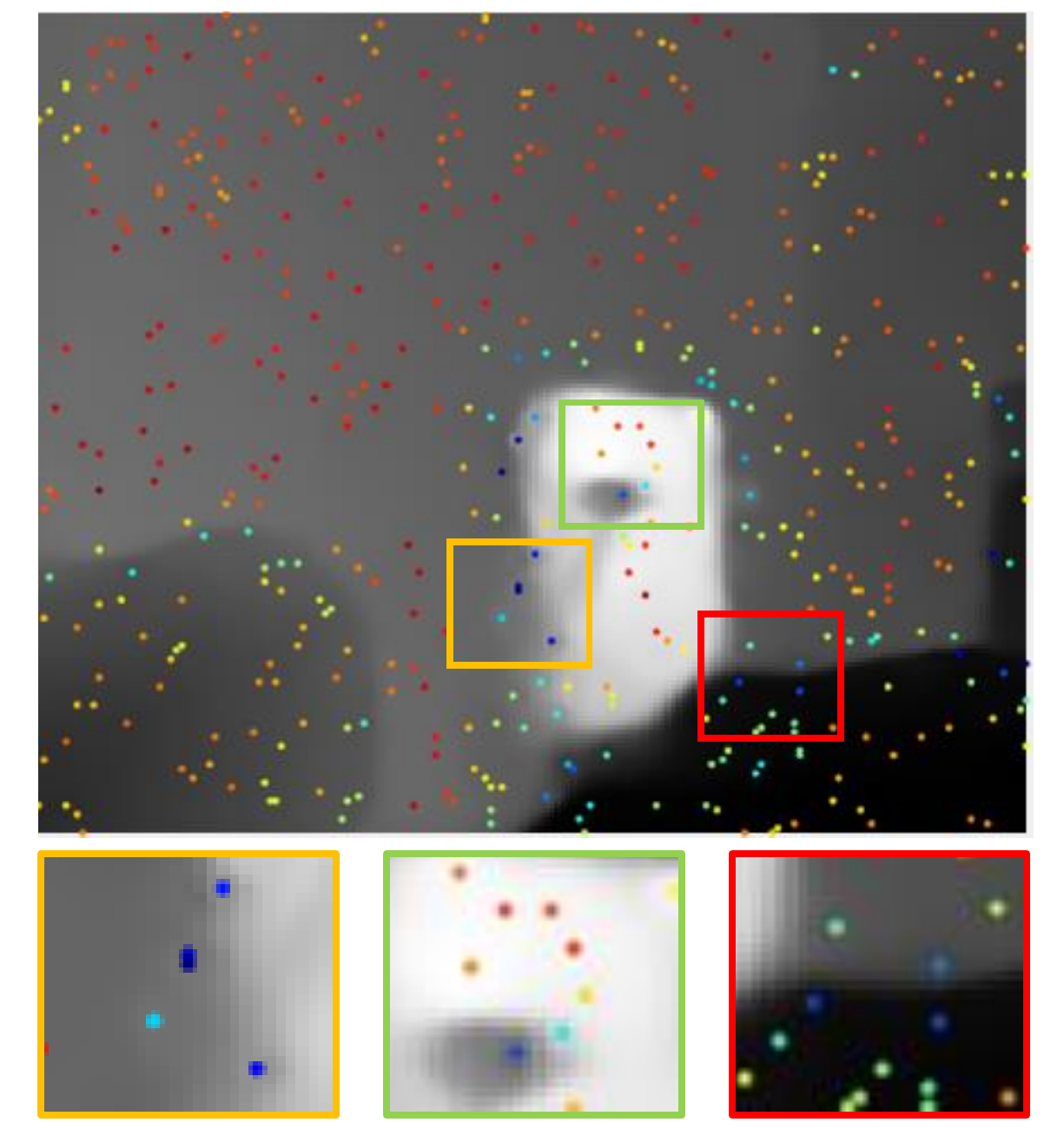} & 
		\includegraphics[width=0.24\columnwidth]{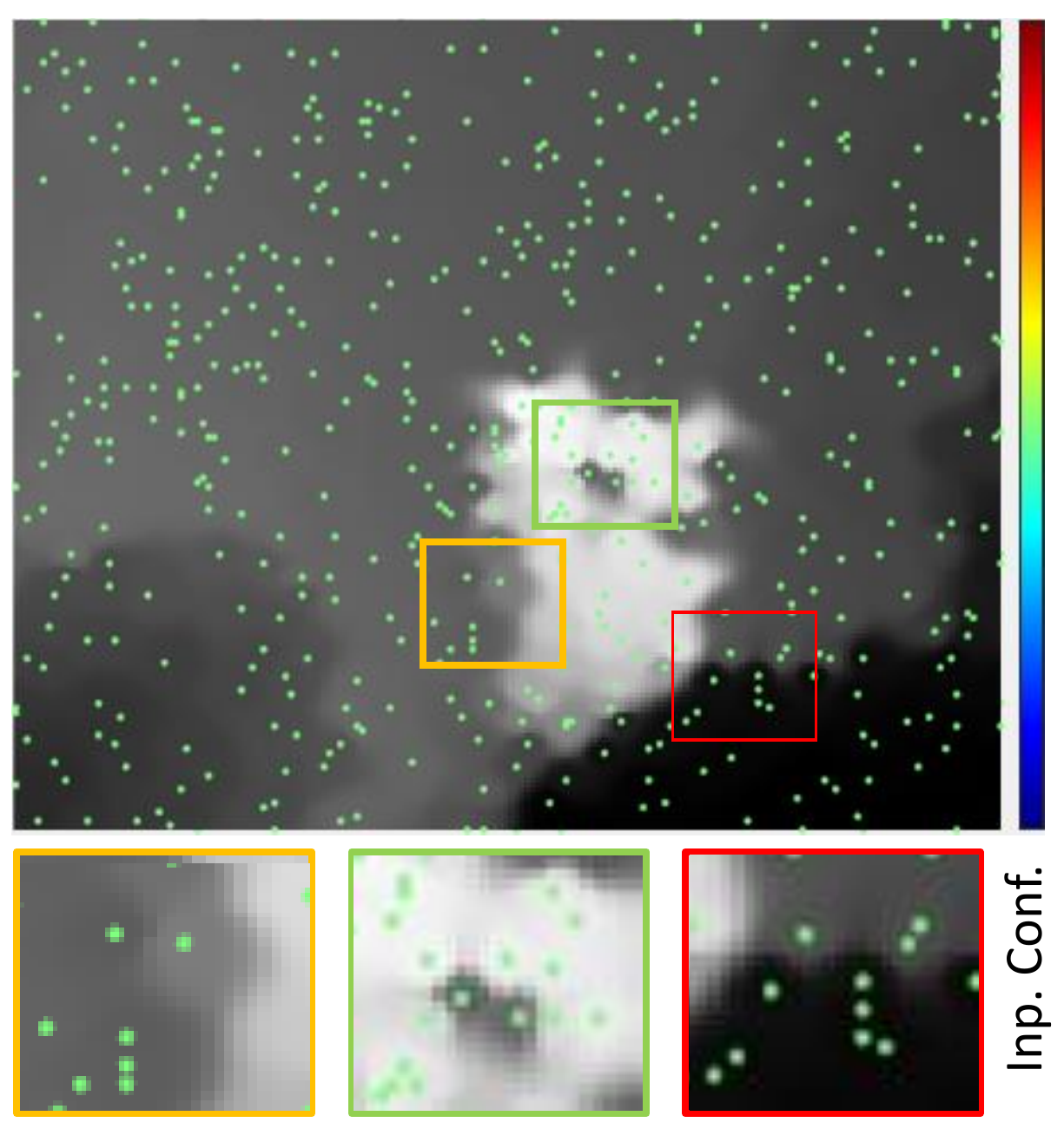} \\ 
	\end{tabular}
	\caption{An examples from the NYU \cite{Silberman12} dataset, where our confidence estimator (left) down-weights depth samples close to edges in order to obtain sharper output. On the other hand, the NCNN \cite{bmvc} struggles along edges due to equally trusting all input samples.}
	\label{fig:nyu_conf}
\end{figure}

Figure \ref{fig:nyu_conf} shows some qualitative examples on the NYU dataset \cite{Silberman12} for our \emph{NCNN-Conf-L1} compared to the standard NCNN \cite{bmvc}. In these examples, the sparse input is \emph{undisturbed} and NCNN should perform well using the binary input confidences. However, NCNN struggles along edges due to equally trusting the background and the foreground. Our \emph{NCNN-Conf-L1} on the other hand, learns proper input confidences that preserve edges similar to non-linear filtering.

\renewcommand{\arraystretch}{0.4} 
\setlength{\tabcolsep}{0.2pt} 
\setlength{\tabcolsep}{5pt} 


\subsection{Sparse Optical Flow Rectification}
We include more results for the sparse optical flow rectification to demonstrate the generalization capabilities of our approach to other types of data. Qualitative examples are shown in Figure \ref{fig:flow_sample1} and \ref{fig:flow_sample2}. Our method successfully removes noisy flow vectors despite the fact that they look completely random. This demonstrates the generalization capabilities of our approach in identifying the inherent noise in the data in a self-supervised manner.

\newcommand{\flowsample}[4]{
	
	\renewcommand{\arraystretch}{0.5} 
	\setlength{\tabcolsep}{0.5pt} 
	\begin{figure}[h]
		\begin{tabular}{C{0.04\textwidth} C{0.48\textwidth} C{0.48\textwidth}}
			\rotatebox{90}{ \ \qquad Grayscale Image} &
			\fbox{\includegraphics[width=0.47\textwidth]{fig/flow/rgb/#1.png}} & 		\fbox{\includegraphics[width=0.47\textwidth]{fig/flow/rgb/#4.png}}\\
			\rotatebox{90}{ \ \ \ Sparse Raw Flow Input} &
			\fbox{\includegraphics[width=0.47\textwidth]{fig/flow/#1_raw_input_flow.png}} & \fbox{\includegraphics[width=0.47\textwidth]{fig/flow/#4_raw_input_flow.png}}\\
			\rotatebox{90}{ \ \ \qquad  Predicted Flow} &
			\fbox{\includegraphics[width=0.47\textwidth]{fig/flow/#1_net_output_flow.png}}& \fbox{\includegraphics[width=0.47\textwidth]{fig/flow/#4_net_output_flow.png}}\\
			\rotatebox{90}{ \quad \qquad  Groundtruth} &
			\fbox{\includegraphics[width=0.47\textwidth]{fig/flow/#1_gt_flow.png}}& \fbox{\includegraphics[width=0.47\textwidth]{fig/flow/#4_gt_flow.png}}\\
			\\
		\end{tabular}
		\caption{ #2	}
		\label{fig:flow_sample#3}
	\end{figure}
}

\flowsample{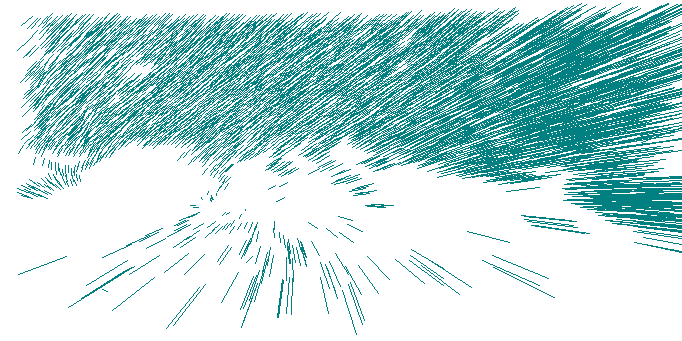}{Two validation samples which highlight the networks noise reduction ability. To the left: tracking failures on the nearly homogeneous road. To the right tracking failures caused by glare. Note that the grayscale image is for visualization and not used.}{1}{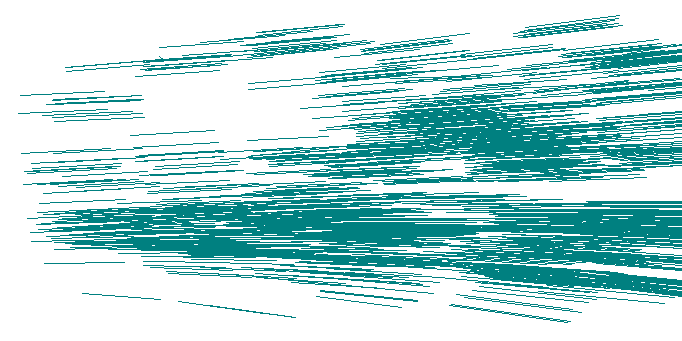}

\flowsample{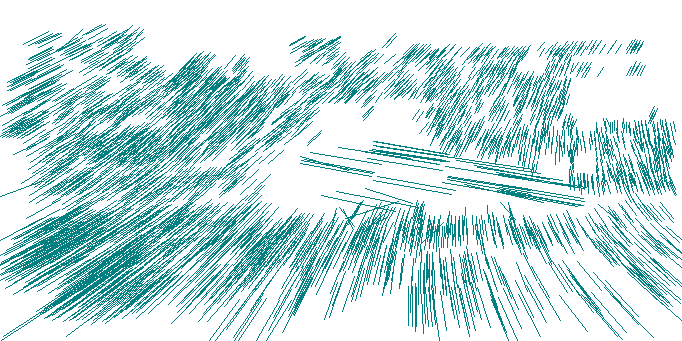}{Left: validation sample with moving rigid objects, demonstrating that the system is not limited to a single epipolar geometry. Right: tracking failure cased by road reflection that is also rectified by our method. Note that the grayscale image is for visualization and not used.}{2}{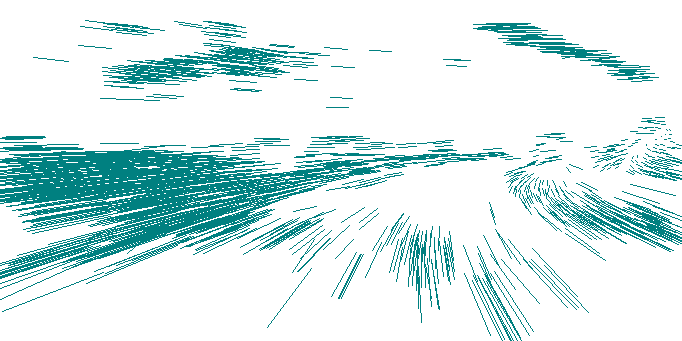}
